\pdfoutput=1

\documentclass{article}
\usepackage[preprint]{neurips_2026}

\usepackage[utf8]{inputenc}
\usepackage[T1]{fontenc}
\usepackage{microtype}
\setcitestyle{authoryear,round,citesep={;},aysep={,},yysep={;}}

\usepackage{amsmath,amsfonts,bm}

\def\eqref#1{equation~\ref{#1}}

\def\1{\bm{1}}

\DeclareMathAlphabet{\mathsfit}{\encodingdefault}{\sfdefault}{m}{sl}
\SetMathAlphabet{\mathsfit}{bold}{\encodingdefault}{\sfdefault}{bx}{n}

\def\gA{{\mathcal{A}}}

\def\gC{{\mathcal{C}}}
\def\gD{{\mathcal{D}}}

\def\gJ{{\mathcal{J}}}

\def\gL{{\mathcal{L}}}
\def\gM{{\mathcal{M}}}
\def\gN{{\mathcal{N}}}

\def\gS{{\mathcal{S}}}

\def\gU{{\mathcal{U}}}

\newcommand{\E}{\mathbb{E}}

\newcommand{\R}{\mathbb{R}}

\DeclareMathOperator*{\argmax}{arg\,max}

\usepackage{graphicx}
\usepackage{booktabs}
\usepackage{amsthm}
\newtheorem{theorem}{Theorem}
\newtheorem{proposition}[theorem]{Proposition}
\newtheorem{lemma}[theorem]{Lemma}
\newtheorem{corollary}[theorem]{Corollary}
\theoremstyle{definition}
\newtheorem{assumption}[theorem]{Assumption}
\theoremstyle{plain}
\newtheorem*{restatedthm}{Theorem~\ref{thm:transport}}
\newtheorem*{restatedprop}{Proposition~\ref{prop:dual}}
\newtheoremstyle{spacedremark}{8pt}{8pt}{\normalfont}{}{\itshape}{.}{.5em}{}
\theoremstyle{spacedremark}
\newtheorem{remark}[theorem]{Remark}
\usepackage{xspace}
\usepackage{algorithm}
\usepackage{algpseudocode}
\usepackage[most]{tcolorbox}
\newtcolorbox{resultbox}{
  enhanced,
  colback=black!4, colframe=black!60,
  boxrule=0pt, leftrule=3pt, arc=0pt,
  left=8pt, right=8pt, top=2pt, bottom=6pt,
  before skip=10pt, after skip=10pt
}
\usepackage[colorlinks=true, allcolors=blue]{hyperref}
\usepackage{url}
\usepackage{placeins}

\makeatletter
\newsavebox{\bqt@box}
\newdimen\bqt@room
\newcommand{\bqt@percent}[2]{\the\numexpr 100*\dimexpr#1\relax/\dimexpr#2\relax\relax\@percentchar}
\newcommand{\bqt@fittabular}{\let\bqt@tabular\tabular \let\bqt@endtabular\endtabular
  \renewenvironment{tabular}{\begin{lrbox}{\bqt@box}\bqt@tabular}{\bqt@endtabular\end{lrbox}\ifdim\wd\bqt@box>\linewidth
      \wlog{arXiv fit: tabular of \@ifundefined{@captype}{the text}{\@captype\space\csname the\@captype\endcsname} scaled to \bqt@percent{\linewidth}{\wd\bqt@box}}\resizebox{\linewidth}{!}{\usebox\bqt@box}\else
      \usebox\bqt@box
    \fi}}
\newcommand{\bqt@fitequation}{\global\let\bqt@lab\@empty \global\let\bqt@tag\@empty
  \setbox\bqt@box\hbox{\def\label##1{\gdef\bqt@lab{##1}}\def\tag##1{\gdef\bqt@tag{##1}}$\displaystyle\BODY$}\begingroup
    \ifx\bqt@tag\@empty \advance\c@equation\@ne \setbox\z@\hbox{\normalfont(\theequation)}\else \setbox\z@\hbox{\normalfont(\bqt@tag)}\fi
    \global\bqt@room=\dimexpr\linewidth-\wd\z@-\fontdimen6\textfont\tw@-1pt\relax
    \gdef\bqt@scale{0}\ifdim\wd\bqt@box>\bqt@room
      \hbadness\@M \hfuzz\maxdimen
      \setbox\z@\hbox to\bqt@room{\unhcopy\bqt@box}\ifnum\badness>\@M \gdef\bqt@scale{1}\fi
    \fi
  \endgroup
  \if1\bqt@scale
    \wlog{arXiv fit: equation \bqt@lab\space scaled to \bqt@percent{\bqt@room}{\wd\bqt@box}}\bqt@equation
      \resizebox{\bqt@room}{!}{\usebox\bqt@box}\ifx\bqt@lab\@empty\else\expandafter\label\expandafter{\bqt@lab}\fi
      \ifx\bqt@tag\@empty\else\expandafter\tag\expandafter{\bqt@tag}\fi
    \bqt@endequation
  \else
    \bqt@equation\BODY\bqt@endequation
  \fi}
\AtBeginDocument{\let\bqt@equation\equation \let\bqt@endequation\endequation
  \RenewEnviron{equation}{\bqt@fitequation}\let\bqt@maketitle\maketitle
  \def\maketitle{\bqt@maketitle\bqt@fittabular}}
\makeatother

\title{Behavior-Flow Q-Gradient Transport: Selection-Free One-Step Flow-Map Policies}

\author{Jinchang Zhou \quad Hanze Jia \quad Jianing Yin \quad Tan Tang \\
  State Key Lab of CAD\&CG, Zhejiang University \\
  \texttt{jinchangzhou@zju.edu.cn}
}

\hypersetup{
  pdftitle={Behavior-Flow Q-Gradient Transport: Selection-Free One-Step Flow-Map Policies},
  pdfauthor={Jinchang Zhou, Hanze Jia, Jianing Yin, Tan Tang},
}

\newcommand{\method}{BQT\xspace}
\newcommand{\vgname}{VGQL\xspace}

\begin{document}

\maketitle

\begin{abstract}
Flow-map policies enable efficient inference through single-step action generation, yet achieving effective critic-guided policy improvement without sacrificing this efficiency remains a challenge in offline-to-online reinforcement learning (RL).
Existing methods rely on best-of-$N$ selection, which requires additional policy and critic forward passes at deployment and undermines the efficiency of one-step generation.
We propose Behavior-Flow Q-Gradient Transport (\method), a novel offline-to-online RL algorithm that builds the regression target of the flow map by transporting the action gradient of the critic along a behavior flow fitted to the data.
This target carries the improvement signal of the critic by itself, so a single action sample suffices for data collection, the TD target and evaluation.
It is derived from deterministic optimal control, and we introduce a surrogate objective that sidesteps the trajectory simulation and backward integration required by adjoint-based methods.
We prove that the surrogate is exact when the behavior velocity is affine, and that its error is in general bounded by a term that decays inversely with the number of flow-map jumps.
On 30 tasks from six OGBench domains, \method outperforms prior one-step methods and is comparable to strong multi-step flow policies on hard tasks. \end{abstract}

\section{Introduction}
\label{sec:introduction}
\begin{figure}[!h]
  \centering
  \includegraphics[width=0.57\textwidth]{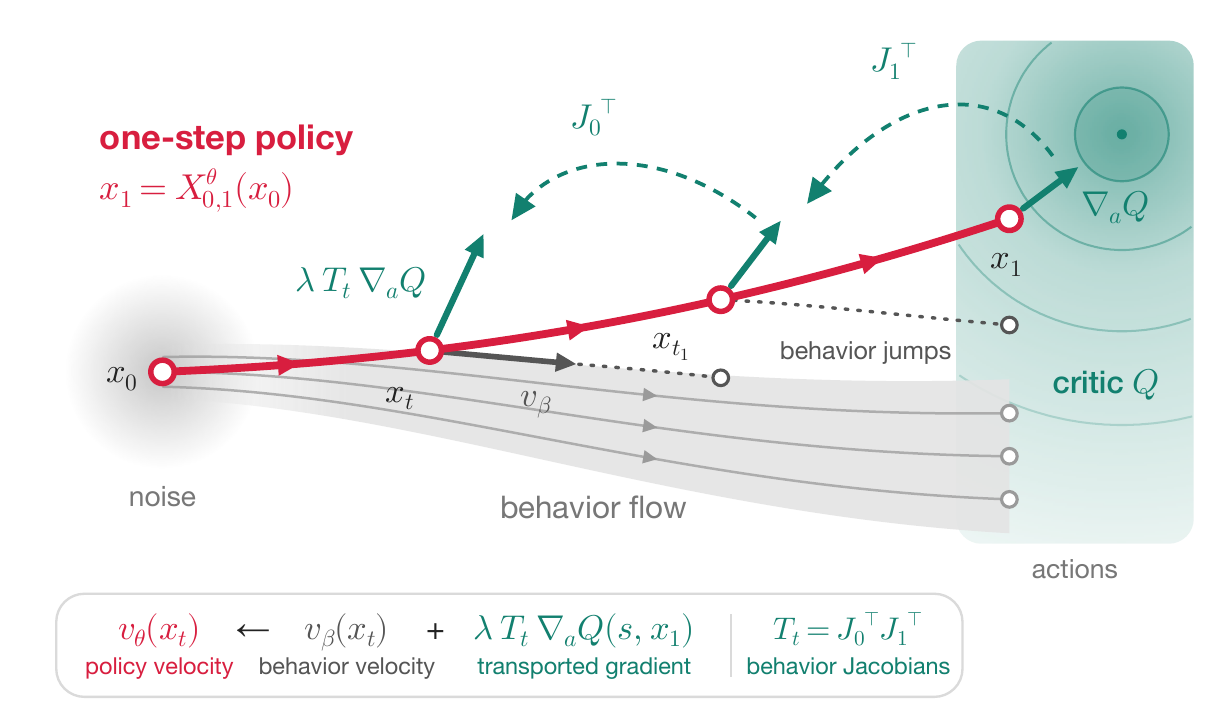}\hfill
  \includegraphics[width=0.41\textwidth]{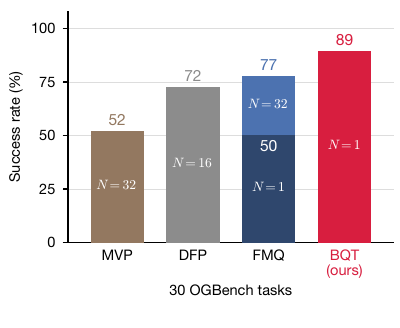}
\caption{\textbf{\method: Behavior-Flow Q-Gradient Transport.} \emph{Left:} \method builds the regression target of a one-step flow-map policy by transporting the action gradient of the critic along the behavior flow. \emph{Right:} Success rate on 30 OGBench tasks. The darker part of FMQ's bar is its success rate with a single sample.}
\label{fig:teaser}
\end{figure}

Diffusion and flow-matching models \citep{ho2020ddpm, song2021scorebased, lipman2023flow, liu2023rectified} are now widely used as policy classes for continuous control. 
Unlike a unimodal Gaussian, they can represent complex, multimodal action distributions, which has made them effective in both imitation learning \citep{chi2023diffusionpolicy} and reinforcement learning (RL) \citep{wang2023diffusion, hansenestruch2023idqlimplicitqlearningactorcritic, park2025flowqlearning, li2026qam}. 
This expressivity, however, comes from iterative generation. 
An action is obtained by integrating a learned denoising process, and every integration step costs one network function evaluation (NFE). 
The resulting latency is paid for every action the policy executes, which motivates policies that retain the expressivity of a flow while generating an action in one or a few NFEs. 
Flow maps \citep{boffi2024flowmap, boffi2025flowmaps, geng2025meanflow} make such policies possible. 
A flow map learns the solution map of the flow-matching ODE directly, so that a single network evaluation carries a noise sample to an action.

Achieving low latency is not enough in offline-to-online RL. 
After pre-training on offline data, the policy must continue to improve under the guidance of a learned critic. 
Recent flow-map policies for this setting bring the critic in mainly through two routes. 
The first is best-of-$N$ selection, a form of policy extraction by rejection sampling \citep{park2025flowqlearning, li2025qchunking}. 
In MVP \citep{zhan2026meanflowpolicy}, the policy draws $N$ candidate actions and the critic selects among them to both construct the learning target and determine the executed action.
DFP \citep{koo2026driftingfieldpolicyonestep}, a one-step policy built on a drifting model \citep{deng2026generativemodelingdrifting}, relies on the same mechanism. 
Alternatively, FMQ \citep{ziakas2026aligningflowmappolicies} pursues the second route by deriving the flow map's regression target from the critic's action gradient, which removes selection from the learning target of the policy.
In our experiments, however, the success rate of FMQ on most manipulation tasks drops substantially when best-of-$N$ selection at test time is replaced by a single sample (Figure~\ref{fig:teaser}).
This reliance on test-time selection undermines the efficiency that one-step generation is meant to provide \citep{zhou2026testtimegradientguidance, zhou2026advantageguidance}. 
Every executed action requires $N$ policy and $N$ critic forward passes rather than a single policy forward pass \citep{park2025flowqlearning}, and the critic can no longer be discarded after training. 
The gap between single-sample and best-of-$N$ performance also shows that part of the improvement comes from the selection step rather than from the flow map.

Our goal is a flow-map policy that requires no selection during training or at test time. 
Without selection, the improvement signal of the critic must be carried entirely by the regression target of the policy. 
We propose Behavior-Flow Q-Gradient Transport (\method), which builds this target by transporting the action gradient of the critic along a behavior flow fitted to the data. 
The target is derived from deterministic optimal control, where policy improvement is posed as steering the behavior flow toward actions of higher critic value at a quadratic control cost. 
For flow maps, WTF \citep{mammadov2026wtfsimulationfreereinforcementlearning} directly matches the value gradient of this control problem, which requires the Jacobian of the policy flow map. 
In contrast, our formulation is rooted in Pontryagin's maximum principle, under which the target is a lean adjoint \citep{domingoenrich2025adjointmatchingfinetuningflow, liu2026valuegradientguidanceflow}. 
On a multi-step flow, computing this adjoint requires simulating a trajectory and integrating an ODE backward along it.
\method replaces both steps with a simulation-free surrogate: jumps of the policy flow map give on-policy intermediate actions, and the action gradient of the critic is transported back to them through the Jacobians of a few behavior flow-map jumps.
The surrogate recovers the lean adjoint exactly when the behavior velocity is affine, in which case fixed points of the regression are stationary points of the control problem. 
In general, we prove an error bound that decays inversely with the number of jumps.
Combined with a standard temporal-difference backup, \method effectively extracts a one-step policy from the critic without any best-of-$N$ selection.

Our main contribution is \method, an offline-to-online RL algorithm that trains a one-step flow-map policy by regression onto a target built from the action gradient of the critic, transported along the behavior flow. 
Unlike prior flow-map policies, \method uses no best-of-$N$ selection, during training or at test time. 
We evaluate \method on 30 tasks from six OGBench domains \citep{park2025ogbench}, where it outperforms prior one-step methods and is comparable to strong multi-step flow policies \citep{li2026qam, dong2026trustregionqadjoint} on hard tasks with headroom.
 
\section{Related Work}
\label{sec:related}

\textbf{Offline-to-online RL.} Offline-to-online RL pre-trains an agent on a fixed dataset and then continues learning with online interaction. A central difficulty is the transition between the two phases. A critic trained only on dataset actions can be inaccurate on the actions the agent takes once it starts to explore, and the resulting updates can degrade the policy obtained offline \citep{lee2022offlineonline, nakamoto2023calql}. Prior work addresses this mainly on the side of the critic and the data, by keeping the pre-trained value function conservative, in-sample or calibrated \citep{kumar2020cql, kostrikov2022iql, nakamoto2023calql}, by constraining the policy to the data with advantage-weighted updates \citep{nair2020awac}, by balancing offline and online samples in the replay buffer \citep{lee2022offlineonline}, or by running an off-policy algorithm on both data sources without a pre-training phase \citep{ball2023rlpd}. More recent work brings expressive diffusion and flow policies to this setting \citep{park2025flowqlearning, li2025qchunking, li2026qam, dong2026trustregionqadjoint}. Following FQL \citep{park2025flowqlearning} and QAM \citep{li2026qam}, we use the same objective in the offline and the online phase. Our policy differs from both: it is a flow map, which needs neither distillation into a separate one-step network nor a multi-step sampler.

\textbf{Policy extraction for generative policies.} Improving a policy from the critic is known as policy extraction \citep{park2025flowqlearning}. For diffusion and flow policies, existing methods use weighted regression \citep{nair2020awac, kang2023efficientdiffusionpoliciesoffline, ding2024diffusionbasedreinforcementlearningqweighted, sochopoulos2026optimaltransportqlearningflow}, best-of-$N$ sampling among candidate actions \citep{hansenestruch2023idqlimplicitqlearningactorcritic, li2025qchunking}, or backpropagation of the critic through the sampling process or through a distilled one-step network \citep{wang2023diffusion, park2025flowqlearning}. A fourth route builds regression targets for the intermediate denoising steps from the action gradient of the critic. Adjoint matching derives them from stochastic optimal control \citep{li2026qam, dong2026trustregionqadjoint, dong2026qlearningscalaradjoint}, and other constructions have been proposed \citep{doo2026qflowstableexpressivereinforcement, wang2026qvgmqguidedvaluegradientmatching}. One-step generative policies follow the same routes. MVP \citep{zhan2026meanflowpolicy} and DFP \citep{koo2026driftingfieldpolicyonestep} select among candidate actions. OFQL \citep{nguyen2026onestepflowql}, MeanFlowQL \citep{wang2026onestepgenerativepolicies} and DriftQL \citep{houssaini2026driftqlearning} backpropagate the critic through the one-step generator, and MeanFlowQL combines this with best-of-$N$ sampling. FMQ \citep{ziakas2026aligningflowmappolicies} builds a regression target for the flow map from the action gradient of the critic, linearized around the behavior action. None of these one-step methods follows the adjoint matching route. We take it for a flow-map policy: the improvement target is derived from deterministic optimal control and carries the improvement by itself, with no selection step.

\textbf{Reward fine-tuning of flow models.} Fine-tuning a pre-trained flow model toward a reward has been formulated in two ways. The first targets the reward-tilted distribution of a KL-regularized problem and is solved by stochastic optimal control \citep{domingoenrich2025adjointmatchingfinetuningflow}. This formulation relies on the stochastic dynamics of a diffusion \citep{mammadov2026wtfsimulationfreereinforcementlearning}, and carrying it over to flow maps requires a stochastic flow map that takes the intermediate time and sample as additional inputs \citep{potaptchik2026metaflowmapsenable}. The second measures the deviation from the pre-trained model by a control cost on the deterministic flow, which gives a deterministic optimal control problem. VGG-Flow \citep{liu2026valuegradientguidanceflow} fine-tunes a multi-step flow by matching a learned value gradient, whereas WTF \citep{mammadov2026wtfsimulationfreereinforcementlearning} fine-tunes a flow map by regressing directly onto the value gradient. \citet{guo2026improvedtechniquesfinetuningflow} give a deterministic version of adjoint matching, which regresses a multi-step flow onto the lean adjoint. OC-Flow \citep{wang2025trainingfreeguidedflow} and FMRG \citep{huang2026guideflowfewstepalignment} solve related deterministic control problems at inference time, without training. In these works, the lean adjoint is integrated backward along a multi-step trajectory, while flow maps are fine-tuned with the value gradient. We target the lean adjoint for a flow-map policy and approximate it with a simulation-free surrogate built from jumps of the policy and behavior flow maps.
 
\section{Preliminaries}
\label{sec:prelim}

\textbf{Offline-to-online RL.} We consider a Markov decision process $\gM = (\gS, \gA, P, r, \gamma)$ with state space $\gS$, action space $\gA \subseteq \R^d$, transition kernel $P(s' \mid s, a)$, reward function $r(s, a)$ and discount factor $\gamma \in [0, 1)$. A policy $\pi(a \mid s)$ is a distribution over actions given the state. The objective is to find a policy that maximizes the expected discounted return $\E_{a_n \sim \pi(\cdot \mid s_n), \; s_{n+1} \sim P(\cdot \mid s_n, a_n)} \big[ \sum_{n=0}^{\infty} \gamma^n \, r(s_n, a_n) \big]$.

In offline-to-online RL, the agent is first pre-trained on a fixed dataset $\gD$ of transitions $(s, a, r, s')$ without access to the environment. It then continues training while interacting with the environment, and the transitions it collects are added to $\gD$. We consider off-policy RL in both phases, where the policy $\pi$ is improved with a critic. It estimates the action-value function, the expected discounted return of taking action $a$ in state $s$ and then following $\pi$: $Q^{\pi}(s, a) = \E_{a_n \sim \pi(\cdot \mid s_n), \; s_{n+1} \sim P(\cdot \mid s_n, a_n)} \big[ \sum_{n=0}^{\infty} \gamma^n \, r(s_n, a_n) \mid s_0 = s, \; a_0 = a \big]$.

\textbf{Flow matching and flow maps.} Flow matching \citep{lipman2023flow, liu2023rectified} generates a sample by integrating the ordinary differential equation (ODE) $\dot{x}_t = b(x_t, t)$ from noise $x_0 \sim \gN(0, I_d)$ at $t = 0$ to $t = 1$, where $b$ is a velocity field. We use it to model the actions in $\gD$: $x_1$ is an action, and $x_t$ with $t < 1$ is an intermediate action. All quantities of the flow are conditioned on the state $s$, which we omit from the notation. The velocity field is trained on the linear interpolant $x_t = (1 - t)\, x_0 + t\, x_1$ between noise $x_0$ and a dataset action $x_1$ by minimizing the conditional flow matching loss
\begin{equation}
  \gL_{\mathrm{CFM}}(\hat{b}) = \E_{(s, x_1) \sim \gD, \; x_0 \sim \gN(0, I_d), \; t \sim \gU[0, 1]} \big\| \hat{b}(x_t, t) - (x_1 - x_0) \big\|^2 .
  \label{eq:cfm}
\end{equation}
The minimizer is $b(x, t) = \E[x_1 - x_0 \mid x_t = x]$, whose flow transports the noise distribution to the distribution of the dataset actions. We call this flow the behavior flow.

Integrating the ODE takes many evaluations of $b$. A flow map \citep{boffi2024flowmap, boffi2025flowmaps} is its solution operator: for $0 \le t \le \tau \le 1$, $X_{t,\tau}$ maps a solution at time $t$ to the same solution at time $\tau$, $X_{t,\tau}(x_t) = x_\tau$, and $X_{0,1}(x_0)$ generates a sample in one step. It is parametrized as
\begin{equation}
  X_{t,\tau}(x) = x + (\tau - t)\, v(x, t, \tau),
  \label{eq:flow_map}
\end{equation}
where $v(x, t, \tau)$ is the average velocity over $[t, \tau]$. On the diagonal $\tau = t$, it equals the instantaneous velocity, $v(x, t, t) = b(x, t)$. The flow map is characterized by any one of three equivalent conditions, the Lagrangian, Eulerian and semigroup conditions, and each of them gives a training objective for $v$ \citep{boffi2025flowmaps}. We use the Eulerian condition $\partial_t X_{t,\tau}(x) + \partial_x X_{t,\tau}(x)\, b(x, t) = 0$, where $\partial_x X_{t,\tau}$ is the Jacobian of the flow map. In terms of the average velocity, it reads
\begin{equation}
  v(x, t, \tau) = b(x, t) + (\tau - t) \big[ \partial_t v(x, t, \tau) + \partial_x v(x, t, \tau)\, b(x, t) \big],
  \label{eq:eulerian_v}
\end{equation}
the identity on which MeanFlow is trained \citep{geng2025meanflow}.

\textbf{Deterministic optimal control.} Deterministic optimal control adds a control $u(x, t)$ to the velocity of the behavior flow and trades a terminal reward against a quadratic cost of the control,
\begin{equation}
  \sup_{u} \; \E_{x_0 \sim \gN(0, I_d)} \Big[ \lambda\, g(x_1^u) - \frac{1}{2} \int_0^1 \big\| u(x_t^u, t) \big\|^2 \, dt \Big]
  \quad \text{s.t.} \quad \dot{x}_t^u = b(x_t^u, t) + u(x_t^u, t), \quad x_0^u = x_0 ,
  \label{eq:oc}
\end{equation}
where $g$ is the terminal reward and the coefficient $\lambda > 0$ scales it against the control cost. Unlike stochastic optimal control \citep{domingoenrich2025adjointmatchingfinetuningflow}, this deterministic formulation carries over to flow maps directly (see \emph{Reward fine-tuning of flow models} in Section~\ref{sec:related}). Its solution transports individual samples toward higher reward instead of reweighting them \citep{mammadov2026wtfsimulationfreereinforcementlearning}.

For a control $u$, let $x_\tau^u$, $\tau \in [t, 1]$, be the trajectory of the controlled flow that starts from $x$ at time $t$. The value function of $u$ is the terminal reward of this trajectory minus its remaining control cost,
\begin{equation}
  V^u(x, t) = \lambda\, g(x_1^u) - \frac{1}{2} \int_t^1 \big\| u(x_\tau^u, \tau) \big\|^2 \, d\tau .
  \label{eq:value}
\end{equation}
The optimal value function $V^*(x, t) = \sup_u V^u(x, t)$ satisfies the Hamilton--Jacobi--Bellman (HJB) equation with terminal condition $V^*(x, 1) = \lambda\, g(x)$, and the optimal control is its gradient,
\begin{equation}
  \partial_t V^*(x, t) + b(x, t)^\top \nabla_x V^*(x, t) + \frac{1}{2} \big\| \nabla_x V^*(x, t) \big\|^2 = 0, \qquad u^*(x, t) = \nabla_x V^*(x, t).
  \label{eq:hjb}
\end{equation}
The optimal velocity field is therefore $b + \nabla_x V^*$, and value-gradient methods fine-tune a flow by regressing its control onto a value gradient \citep{liu2026valuegradientguidanceflow, mammadov2026wtfsimulationfreereinforcementlearning}.
 
\section{Behavior-Flow Q-Gradient Transport}
\label{sec:method}

\subsection{Behavior flow-map policy}
\label{sec:behavior}

We model the actions in $\gD$ with a behavior flow map $X^\beta_{t,\tau}(x) = x + (\tau - t)\, v_\beta(x, t, \tau)$, whose average velocity $v_\beta$ is a network with parameters $\beta$. It is trained with two losses. The first one, $\gL_{\mathrm{diag}}(\beta)$, is the conditional flow matching loss of \eqref{eq:cfm} with $\hat{b}(x, t) = v_\beta(x, t, t)$. The second one is the MeanFlow loss \citep{geng2025meanflow}, which enforces the Eulerian condition in the form of \eqref{eq:eulerian_v} off the diagonal,
\begin{equation}
  \gL_{\mathrm{cons}}(\beta) = \E \Big\| v_\beta(x_t, t, \tau) - \mathrm{sg} \Big[ (x_1 - x_0) + (\tau - t) \big[ \partial_t v_\beta(x_t, t, \tau) + \partial_x v_\beta(x_t, t, \tau)\, (x_1 - x_0) \big] \Big] \Big\|^2 ,
  \label{eq:cons}
\end{equation}
where $\mathrm{sg}[\cdot]$ is the stop-gradient operator, $(t, \tau)$ is uniform on $0 \le t \le \tau \le 1$, and $s$, $x_1$ and $x_0$ are sampled as in \eqref{eq:cfm}. The velocity $b$ of \eqref{eq:eulerian_v} is replaced by the conditional velocity $x_1 - x_0$ of the interpolant, as in \eqref{eq:cfm}. The behavior flow map minimizes $\gL_{\mathrm{diag}}(\beta) + \gL_{\mathrm{cons}}(\beta)$.

The policy that acts in the environment is a second flow map $X^\theta_{t,\tau}(x) = x + (\tau - t)\, v_\theta(x, t, \tau)$, which generates an action in one step, $a = X^\theta_{0,1}(x_0)$ with $x_0 \sim \gN(0, I_d)$. Its average velocity $v_\theta$ is initialized with the parameters of $v_\beta$, so that the two flow maps coincide at the start of training.

\subsection{Surrogate objective}
\label{sec:surrogate}

\textbf{Lean adjoint matching.} For each state $s$, we instantiate the control problem of \eqref{eq:oc} with the critic $g = Q(s, \cdot)$ as the terminal reward and with the behavior flow of Section~\ref{sec:behavior} as the uncontrolled flow, so from here on $b(x, t) = v_\beta(x, t, t)$. The control of the policy is then the difference between its diagonal velocity and the behavior velocity, $u(x, t) = v_\theta(x, t, t) - b(x, t)$. Besides its HJB characterization in \eqref{eq:hjb}, the optimal control is characterized by Pontryagin's maximum principle (PMP) \citep{liu2026valuegradientguidanceflow}. Let $x_\tau$, $\tau \in [t, 1]$, denote the trajectory through $x_t$ of the flow controlled by $u^*$. Then $u^*(x_t, t) = \tilde{p}_t$, where $\tilde{p}$ solves
\begin{equation}
  \frac{d}{d\tau}\, \tilde{p}_\tau = - \partial_x b(x_\tau, \tau)^\top \tilde{p}_\tau, \qquad \tilde{p}_1 = \lambda\, \nabla_a Q(s, x_1).
  \label{eq:lean}
\end{equation}
In adjoint matching \citep{domingoenrich2025adjointmatchingfinetuningflow}, $\tilde{p}$ is called the lean adjoint, and this condition is turned into a regression for the control of the current policy,
\begin{equation}
  \gL_{\mathrm{adj}}(\theta) = \E\, \big\| u(x_t, t) - \mathrm{sg} [ \tilde{p}_t ] \big\|^2 ,
  \label{eq:lam}
\end{equation}
where $x_t$ lies on a trajectory of the current policy, along which $\tilde{p}$ solves \eqref{eq:lean}. This regression has the same fixed points as matching the value gradient of the current policy, namely the stationary points of the control problem in \eqref{eq:oc} \citep[and Appendix~\ref{app:fixed_point}]{liu2026valuegradientguidanceflow}.

\textbf{Simulation-free transport.} The two targets differ in what they require. The value gradient of the current policy involves the Jacobian of the policy. The lean adjoint does not, but it requires two operations: simulating a trajectory of the policy, and integrating \eqref{eq:lean} backward along it. \method replaces the simulation with jumps of the policy flow map, and the backward integration with Jacobians of the behavior flow map. For the trajectory, a jump of the policy flow map from a noise sample $x_0$ gives the on-policy intermediate action $x_t = \mathrm{sg}[X^\theta_{0,t}(x_0)]$, and $K$ further jumps $x_{t_{k+1}} = X^\theta_{t_k, t_{k+1}}(x_{t_k})$ over the knots $t = t_0 < t_1 < \dots < t_K = 1$ give the rest of the trajectory.

\begin{table}[b]
\caption{Guidance terms added to the behavior velocity $v_\beta(x_t, t, t)$ in the regression target of the diagonal velocity of the policy. $\hat{C}_t(x_t) = \frac{1 - t}{2} \| u(x_\tau, \tau) \|^2$, with $\tau \sim \gU[t, 1]$ and $x_\tau = X^\theta_{t,\tau}(x_t)$, is a one-sample estimate of the control cost of the policy on $[t, 1]$.}
\label{tab:targets}
\begin{center}
\begin{tabular}{lll}
\toprule
Target & Guidance term & Computed with \\
\midrule
Value gradient & $\nabla_{x_t} \big[ \lambda\, Q \big( s, X^\theta_{t,1}(x_t) \big) - \hat{C}_t(x_t) \big]$ & Jacobian of $X^\theta$ \\
Value gradient (Euclidean) & $\lambda\, \nabla_a Q(s, x_1) - \nabla_{x_t} \hat{C}_t(x_t)$ & Jacobian of $X^\theta$ \\
Lean adjoint & $\lambda\, T^\star_t\, \nabla_a Q(s, x_1)$ & numerical integration of \eqref{eq:lean} \\
Identity transport & $\lambda\, \nabla_a Q(s, x_1)$ & no Jacobian product \\
\method (ours) & $\lambda\, T_t\, \nabla_a Q(s, x_1)$ & $K$ Jacobian products of $X^\beta$ \\
\bottomrule
\end{tabular}
\end{center}
\end{table}

It remains to replace the backward integration. Since \eqref{eq:lean} is linear, the lean adjoint is a linear function of its terminal value, $\tilde{p}_t = \lambda\, T^\star_t\, \nabla_a Q(s, x_1)$, where the operator $T^\star_t$ transports the action gradient of the critic from $x_1$ back to $x_t$. Along a trajectory of the behavior flow, the transport over $[t_k, t_{k+1}]$ is the transposed Jacobian of the jump $X^\beta_{t_k, t_{k+1}}$, which requires no integration. We therefore approximate $T^\star_t$ by chaining $K$ such Jacobians, each evaluated at a point of the policy trajectory,
\begin{equation}
  T_t = J_0^\top J_1^\top \cdots J_{K-1}^\top, \qquad J_k = \partial_x X^\beta_{t_k, t_{k+1}}(x_{t_k}) .
  \label{eq:transport}
\end{equation}
Applying $T_t$ to the action gradient takes $K$ vector-Jacobian products. The policy minimizes
\begin{equation}
  \gL_{\mathrm{guide}}(\theta) = \E_{s \sim \gD, \; x_0 \sim \gN(0, I_d), \; t \sim \gU[0, 1]} \Big\| v_\theta(x_t, t, t) - \mathrm{sg} \big[ v_\beta(x_t, t, t) + \lambda\, T_t\, \nabla_a Q(s, x_1) \big] \Big\|^2 ,
  \label{eq:guide}
\end{equation}
a surrogate objective for lean adjoint matching. Theorem~\ref{thm:transport} bounds the error of replacing $T^\star_t$ by $T_t$.

\begin{resultbox}
\begin{theorem}
\label{thm:transport}
Under the regularity conditions of Appendix~\ref{app:proofs}, assume that $X^\beta$ and $X^\theta$ satisfy the flow-map conditions of Section~\ref{sec:prelim}, that is, they are the flow maps of the velocities $b$ and $b + u$. Assume further that, for all $x$ and $\tau$, the eigenvalues of the symmetric part of $\partial_x b(x, \tau)$ are at most $\mu$ and $\partial_x b(\cdot, \tau)$ is $L_2$-Lipschitz. Then, for every $t$ and $x_t$ and for $K$ evenly spaced knots with spacing $h = (1 - t) / K$,
\begin{equation}
  \big\| T_t - T^\star_t \big\| \;\le\; \frac{L_2\, (1 - t)^2}{2 K}\; e^{\mu (1 - t) + \max(\mu, 0)\, h}\, \sup_{\tau \in [t, 1]} \big\| u(x_\tau, \tau) \big\| ,
  \label{eq:transport_bound}
\end{equation}
where $x_\tau = X^\theta_{t,\tau}(x_t)$ is the trajectory of the policy through $x_t$ and the norm on the left is the operator norm. In particular, $T_t = T^\star_t$ for every $K$ if $b$ is affine in $x$.
\end{theorem}
\end{resultbox}

The error of the target is thus of order $1 / K$ and proportional to the magnitude of the control. In the affine case, a fixed point of $\gL_{\mathrm{guide}}$ is a stationary point of the control problem in \eqref{eq:oc}. The proof is given in Appendix~\ref{app:transport}, where Corollary~\ref{cor:residual} bounds the residual of the stationarity condition in the general case.

\textbf{Self-distillation.} The loss $\gL_{\mathrm{guide}}$ trains only the diagonal velocity $v_\theta(x, t, t)$, whereas actions are generated by the one-step map $X^\theta_{0,1}$. Following WTF \citep{mammadov2026wtfsimulationfreereinforcementlearning}, we add a self-distillation loss that keeps $X^\theta$ a valid flow map of its diagonal velocity, as Theorem~\ref{thm:transport} assumes,
\begin{equation}
  \gL_{\mathrm{dist}}(\theta) = \E \Big\| v_\theta(x_t, t, \tau) - \mathrm{sg} \Big[ \bar{v}_t + (\tau - t) \big[ \partial_t v_\theta(x_t, t, \tau) + \partial_x v_\theta(x_t, t, \tau)\, \bar{v}_t \big] \Big] \Big\|^2 ,
  \label{eq:dist}
\end{equation}
where $\bar{v}_t = v_\theta(x_t, t, t)$ is the diagonal velocity at the on-policy intermediate action $x_t = \mathrm{sg}[X^\theta_{0,t}(x_0)]$, and $s$, $x_0$ and $(t, \tau)$ are sampled as in \eqref{eq:cons}.

\textbf{Related targets.} Table~\ref{tab:targets} compares the target of \eqref{eq:guide} with related targets of the same additive form: value-gradient matching as in WTF \citep{mammadov2026wtfsimulationfreereinforcementlearning} and its variant with the Euclidean reward gradient, lean adjoint matching \citep{domingoenrich2025adjointmatchingfinetuningflow, liu2026valuegradientguidanceflow}, and identity transport ($T_t = I$). FMQ \citep{ziakas2026aligningflowmappolicies} shares the idea of identity transport in a different form: it regresses the average velocity $v_\theta(x_t, t, 1)$ at an interpolated $x_t$ onto $v_\beta(x_t, t, 1) + \kappa\, \nabla_a Q(s, a^\beta) / \| \nabla_a Q(s, a^\beta) \|$. Its action gradient is taken at the behavior action $a^\beta = X^\beta_{t,1}(x_t)$ instead of the action of the policy.

\subsection{Control cost budget and practical algorithm}
\label{sec:algorithm}

\textbf{Control cost budget.} Following SAC \citep{haarnoja2018sacapplications} and TRQAM \citep{dong2026trustregionqadjoint}, we replace the choice of $\lambda$ by a budget on the control cost of the policy and adapt $\lambda$ with a dual update. Let $R(u) = \E_{s \sim \gD, \; x_0 \sim \gN(0, I_d)} \big[ Q(s, x_1^u) \big]$ be the expected critic value and $\gC(u) = \E_{s \sim \gD, \; x_0 \sim \gN(0, I_d)} \big[ \frac{1}{2} \int_0^1 \| u(x_t^u, t) \|^2 \, dt \big]$ the expected control cost of a control~$u$. The objective of \eqref{eq:oc} with $g = Q(s, \cdot)$, averaged over the states, is then $\gJ_\lambda(u) = \lambda\, R(u) - \gC(u)$. For a budget $\delta > 0$, we consider the constrained problem
\begin{equation}
  \sup_{u} \; R(u) \quad \text{s.t.} \quad \gC(u) \le \delta ,
  \label{eq:budget}
\end{equation}
with the Lagrangian $\ell(u, \rho) = R(u) - \rho\, \big( \gC(u) - \delta \big)$, the multiplier $\rho \ge 0$ and the dual function $D(\rho) = \sup_u \ell(u, \rho)$. The following proposition, proved in Appendix~\ref{app:dual}, relates this problem to the control problem in \eqref{eq:oc}.

\begin{resultbox}
\begin{proposition}
\label{prop:dual}
\textit{(i)} For every $\rho > 0$ and every control $u$, $\ell(u, \rho) = \rho\, \delta + \rho\, \gJ_{1/\rho}(u)$. Hence $u$ maximizes $\ell(\cdot, \rho)$ if and only if it maximizes $\gJ_\lambda$ with $\lambda = 1 / \rho$, and such a control solves \eqref{eq:budget} if its control cost equals $\delta$.
\textit{(ii)} The dual function $D$ is convex, and $D(\rho)$ is an upper bound on the value of \eqref{eq:budget} for every $\rho \ge 0$. If $u_\rho$ maximizes $\ell(\cdot, \rho)$, then $\delta - \gC(u_\rho)$ is a subgradient of $D$ at $\rho$.
\end{proposition}
\end{resultbox}

By \textit{(ii)}, subgradient descent on the dual function moves $\rho$ in the direction of $\gC(u_\rho) - \delta$, so that $\lambda = 1 / \rho$ decreases when the control cost of $u_\rho$ exceeds the budget and increases otherwise. As in SAC, we take one step in this direction per update of the policy, with the control cost of the current policy in place of that of $u_\rho$,
\begin{equation}
  \log \lambda \;\leftarrow\; \log \lambda - \eta\, \Big( \frac{\hat{\gC}}{\delta} - 1 \Big) ,
  \qquad
  \hat{\gC} = \E_{s \sim \gD, \; x_0 \sim \gN(0, I_d), \; t \sim \gU[0, 1]} \Big[ \frac{1}{2} \big\| v_\theta(x_t, t, t) - v_\beta(x_t, t, t) \big\|^2 \Big] ,
  \label{eq:dual}
\end{equation}
where $\eta$ is the step size. The step acts on $\log \lambda$ and on the relative error, which keeps $\lambda$ positive and makes $\eta$ independent of the scale of $\delta$. After each step, $\lambda$ is clipped to an interval $[\lambda_{\min}, \lambda_{\max}]$. The estimate $\hat{\gC}$ is computed from the intermediate actions $x_t$ and the two velocities of the loss of \eqref{eq:guide}, so the update of $\lambda$ requires no additional network evaluation.

\textbf{Practical algorithm.} The critic $Q_\phi$ is trained by temporal-difference learning,
\begin{equation}
  \gL_{\mathrm{TD}}(\phi) = \E_{(s, a, r, s') \sim \gD, \; x_0 \sim \gN(0, I_d)} \Big( Q_\phi(s, a) - r - \gamma\, Q_{\bar{\phi}} \big( s', X^\theta_{0,1}(x_0) \big) \Big)^2 ,
  \label{eq:td}
\end{equation}
where the next action is the action of the policy at $s'$, and $\bar{\phi}$ is an exponential moving average of $\phi$. Algorithm~\ref{alg:bqt} summarizes \method; we use $K = 2$ evenly spaced jumps. The update is the same in the offline and in the online phase, so the behavior flow map keeps training as the dataset grows. Acting takes a single evaluation of $v_\theta$, with no selection among candidate actions.

\begin{algorithm}[t]
\caption{\method}
\label{alg:bqt}
\begin{algorithmic}[1]
\Require dataset $\gD$, control cost budget $\delta$, number of jumps $K$, step size $\eta$
\State Initialize $\beta$, $\phi$ and $\lambda$; set $\theta \leftarrow \beta$ and $\bar{\phi} \leftarrow \phi$
\For{each update}
  \State \textbf{if} online: execute $a = X^\theta_{0,1}(x_0)$, $x_0 \sim \gN(0, I_d)$, and add the transition to $\gD$
  \State Sample a batch of transitions $(s, a, r, s') \sim \gD$
  \State With $x_0 \sim \gN(0, I_d)$ and $t \sim \gU[0, 1]$: $x_t \leftarrow \mathrm{sg}[X^\theta_{0,t}(x_0)]$, $x_{t_{k+1}} \leftarrow X^\theta_{t_k, t_{k+1}}(x_{t_k})$ for $k = 0, \dots, K - 1$
  \State Compute $T_t\, \nabla_a Q(s, x_1)$ with $K$ vector-Jacobian products of the behavior flow map \Comment{\eqref{eq:transport}}
  \State Gradient step on $\gL_{\mathrm{diag}}(\beta) + \gL_{\mathrm{cons}}(\beta)$, on $\gL_{\mathrm{guide}}(\theta) + \gL_{\mathrm{dist}}(\theta)$ and on $\gL_{\mathrm{TD}}(\phi)$
  \State Update $\lambda$ with \eqref{eq:dual} and $\bar{\phi}$ as a moving average of $\phi$
\EndFor
\end{algorithmic}
\end{algorithm}
 
\section{Experiments}
\label{sec:experiments}

We conduct experiments on a range of challenging long-horizon, sparse-reward tasks to evaluate the effectiveness of \method in offline-to-online RL.

\textbf{Domains and datasets.}
We consider six OGBench domains \citep{park2025ogbench} with five tasks each.
In the four manipulation domains, a robot arm has to rearrange two to four cubes (\texttt{cube-double}, \texttt{cube-triple}, \texttt{cube-quadruple}) or solve a button puzzle (\texttt{puzzle-4x4}).
In the two locomotion domains, an ant (\texttt{antmaze-giant}) or a humanoid (\texttt{humanoidmaze-large}) has to navigate a maze.
The locomotion domains use the default \texttt{navigate} datasets, and \texttt{cube-double} and \texttt{cube-triple} use the default \texttt{play} datasets.
As in QAM \citep{li2026qam}, \texttt{puzzle-4x4} uses the 100M-transition dataset of \citet{park2025horizonreduction}.
For \texttt{cube-quadruple}, we use the first 25M transitions of its 100M-transition dataset.
In the manipulation domains, we follow QC \citep{li2025qchunking} and learn action chunking policies with a chunk size of 5.
Appendix~\ref{app:benchmark} gives more details on the domains.

\textbf{Comparisons.}
We compare \method with four one-step baselines.
MVP \citep{zhan2026meanflowpolicy} and DFP \citep{koo2026driftingfieldpolicyonestep} rely on best-of-$N$ selection, in which the critic selects among $N$ candidate actions drawn from the policy.
FMQ \citep{ziakas2026aligningflowmappolicies} builds the regression target of the flow map from the action gradient of the critic.
\vgname is a baseline built on WTF \citep{mammadov2026wtfsimulationfreereinforcementlearning}.
It regresses the policy flow map directly onto the value gradient of Table~\ref{tab:targets}.
The ablation compares four guidance terms of Table~\ref{tab:targets} within Algorithm~\ref{alg:bqt}: the value gradient (\vgname), its Euclidean variant (\vgname (Euclidean)), identity transport and \method.
Appendix~\ref{app:baselines} gives the details of each baseline.

All methods follow the offline-to-online protocol of QC, with two critics, 1M gradient steps on the offline dataset and then 1M environment steps online.
Each method is run with 8 seeds on every task.
\method and \vgname draw a single action sample in data collection, in the TD target and at evaluation.
MVP, FMQ and DFP select the best of $N$ samples in all three places, with $N = 32$ for MVP and FMQ and $N = 16$ for DFP.
We report the success rate over 50 evaluation episodes at 2M steps, with 95\% bootstrap confidence intervals.
Appendix~\ref{app:hyperparameters} lists the hyperparameters.
 
\section{Results}
\label{sec:results}

\begin{figure}[t]
  \centering
  \includegraphics[width=\textwidth]{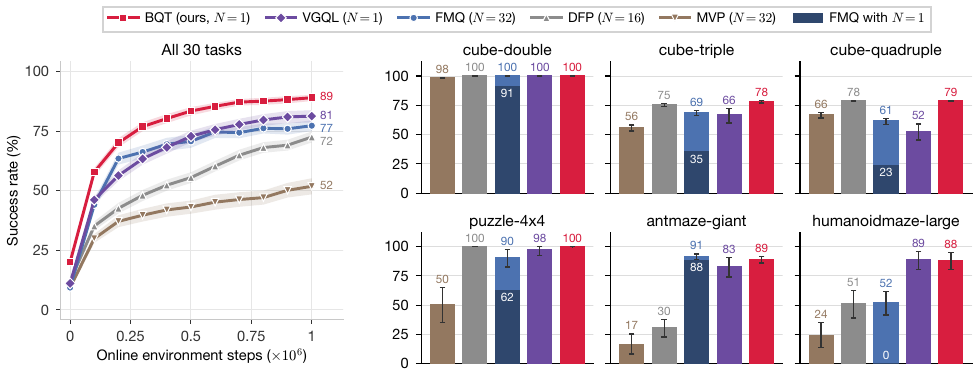}
  \caption{\textbf{Main results (30 tasks, 8 seeds).} \emph{Left:} Success rate averaged over all 30 tasks during online training; step 0 is the offline policy. \emph{Right:} Success rate of each domain at 2M steps. The darker part of FMQ's bars is its success rate with a single sample. Means with 95\% bootstrap confidence intervals.}
  \label{fig:main}
\end{figure}

\begin{figure}[t]
  \centering
  \includegraphics[width=\textwidth]{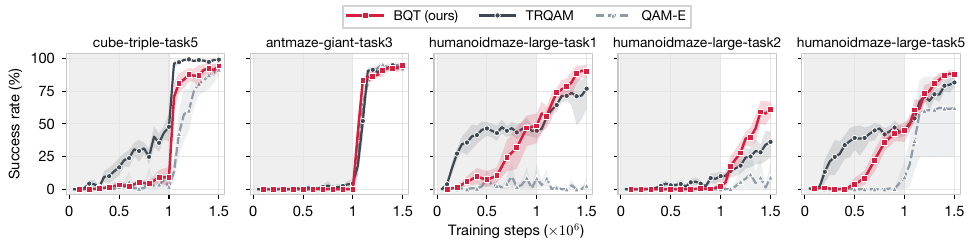}
  \caption{\textbf{Comparison with multi-step flow policies (8 seeds).} Offline phase (shaded) and online phase under the protocol of TRQAM. Means with 95\% bootstrap confidence intervals.}
  \label{fig:multistep}
\end{figure}

\begin{figure}[t]
  \centering
  \includegraphics[width=\textwidth]{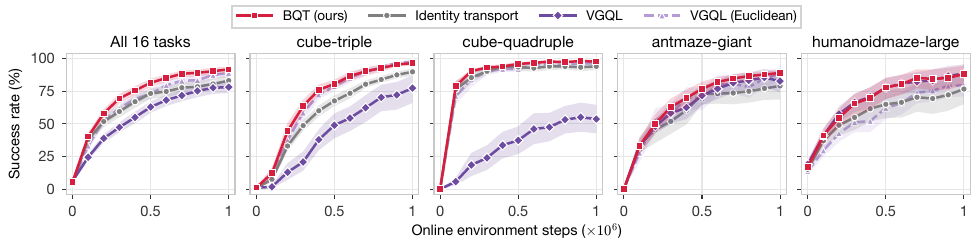}
  \caption{\textbf{Guidance-term ablation (16 tasks, 8 seeds).} Online phase; step 0 is the offline policy. Means with 95\% bootstrap confidence intervals.}
  \label{fig:abl_target}
\end{figure}

\begin{table}[t]
\caption{\textbf{Guidance-term ablation, final success rate (\%).} Means over 8 seeds at 2M steps with 95\% bootstrap confidence intervals. Bold: at or above 95\% of the best value in the column.}
\label{tab:abl_target}
\begin{center}
\small
\setlength{\tabcolsep}{4pt}
\begin{tabular}{lccccc}
\toprule
Target & \texttt{cube-triple} & \texttt{cube-quadruple} & \texttt{antmaze-giant} & \texttt{humanoidmaze-large} & All 16 tasks \\
 & (3 tasks) & (3 tasks) & (5 tasks) & (5 tasks) & \\
\midrule
\vgname & 77.4 {\tiny [66.0, 87.1]} & 53.8 {\tiny [42.6, 65.3]} & 82.8 {\tiny [74.2, 90.2]} & \textbf{88.8} {\tiny [80.7, 95.6]} & 78.2 {\tiny [73.4, 82.6]} \\
\vgname (Euclidean) & \textbf{97.8} {\tiny [96.7, 98.8]} & \textbf{96.2} {\tiny [93.3, 98.2]} & \textbf{88.8} {\tiny [82.7, 93.2]} & 79.6 {\tiny [71.1, 87.6]} & \textbf{89.0} {\tiny [85.8, 91.9]} \\
Identity transport & 89.9 {\tiny [87.1, 92.5]} & \textbf{94.1} {\tiny [91.2, 96.8]} & 79.2 {\tiny [68.9, 88.0]} & 76.8 {\tiny [65.1, 87.7]} & 83.2 {\tiny [78.5, 87.5]} \\
\method (ours) & \textbf{96.6} {\tiny [94.9, 98.1]} & \textbf{97.8} {\tiny [96.5, 98.8]} & \textbf{89.1} {\tiny [86.1, 91.5]} & \textbf{88.2} {\tiny [80.3, 94.9]} & \textbf{91.8} {\tiny [89.3, 94.0]} \\
\bottomrule
\end{tabular}
 \end{center}
\end{table}

\begin{figure}[t]
  \centering
  \includegraphics[width=\textwidth]{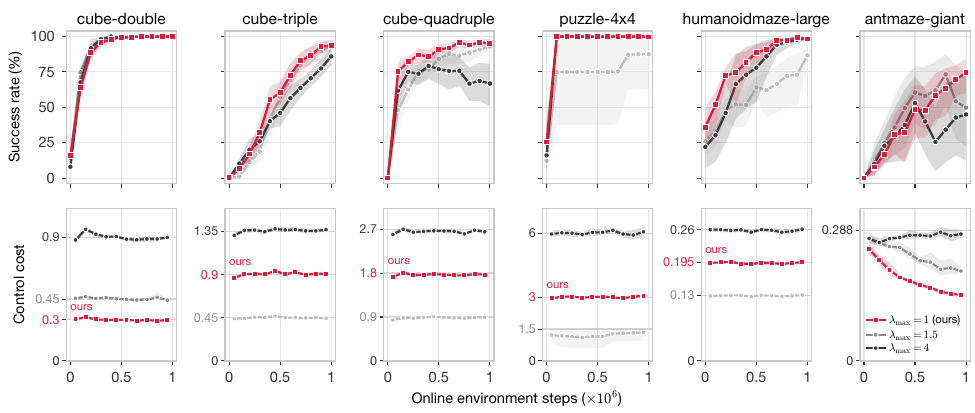}
  \caption{\textbf{Control-cost budget (one task per domain, 8 seeds).} Top: success rate. Bottom: control cost of the policy during online training, with vertical ticks at the three compared budgets $\delta$. The setting of the main results is shown in red. On \texttt{antmaze-giant}, the budget is fixed and the limit $\lambda_{\max}$ is varied.}
  \label{fig:abl_budget}
\end{figure}

\textbf{(Q1) How does \method compare with prior one-step methods?}
Figure~\ref{fig:main} shows the success rate averaged over all 30 tasks during online training and the success rate of each domain at 2M steps.
\method achieves the best overall success rate (89.0\%), ahead of \vgname (81.3\%) and FMQ (77.2\%).
It is also more sample-efficient overall, with the highest average success rate at every evaluation of the online phase.
It reaches an average success rate of 80\% after 0.4M online steps, while \vgname needs 0.9M steps and the other baselines do not reach it.
\method obtains these results with a single action sample.
When best-of-$N$ selection at test time is replaced by a single sample, the success rate of FMQ drops from 77.2\% to 50.0\%.
With the networks of \method, best-of-32 selection requires 89 times the floating-point operations of a single-sample query (Appendix~\ref{app:compute}).
Appendix~\ref{app:tasks} gives the results per domain and per task, with per-task training curves.

\textbf{(Q2) Is \method comparable to strong multi-step flow policies?}
We evaluate \method under the protocol of TRQAM \citep{dong2026trustregionqadjoint} and compare it with the released results of TRQAM and QAM-E \citep{li2026qam}.
We select the six tasks with headroom, on which the final success rate of \method in our main experiments or that of TRQAM is below 85\%.
One of them, \texttt{cube-quadruple-task5}, remains unsolved by all three methods and is not shown.
Figure~\ref{fig:multistep} shows the success rates on the other five tasks over the offline and online phases.
At the end of training, \method is comparable to TRQAM and reaches a higher success rate than QAM-E on all five tasks.
In the offline phase, \method improves more slowly than TRQAM on \texttt{cube-triple-task5}, and it closes most of this gap online.
Appendix~\ref{app:multistep} gives the protocol and final results.

\textbf{(Q3) How does the guidance term affect performance?}
We compare the four guidance terms on 16 tasks: \texttt{task2} to \texttt{task4} of \texttt{cube-triple} and \texttt{cube-quadruple}, and all five tasks of \texttt{antmaze-giant} and \texttt{humanoidmaze-large}.
Figure~\ref{fig:abl_target} shows the success rates during online training, and Table~\ref{tab:abl_target} reports them at 2M steps.
\method achieves the best overall success rate (91.8\%) and is the most robust target across domains.
For example, \vgname is the best target on \texttt{humanoidmaze-large}, but it has the lowest success rate on \texttt{cube-triple} and \texttt{cube-quadruple}.
Its Euclidean variant performs better overall.
Identity transport differs from \method only in $T_t$, and neither loss contains a control cost term.
Replacing the Jacobians of the behavior flow map by the identity lowers the success rate in every domain.
Per-task results are in Appendix~\ref{app:ablation}.

\textbf{(Q4) How sensitive is \method to the control cost budget?}
For each domain, we select one representative task and compare three budgets $\delta$ (Figure~\ref{fig:abl_budget}).
On \texttt{antmaze-giant}, we found it better to limit $\lambda$ directly, so this panel compares three values of $\lambda_{\max}$ at the same budget.
In the other five domains, the dual update of \eqref{eq:dual} keeps the control cost close to its budget throughout the online phase.
We choose a small budget among those that learn fast and reach a high final success rate, since a smaller budget keeps the policy closer to the behavior flow.
Overall, the budget affects the success rate and has to be tuned for each domain.
Table~\ref{tab:abl_budget} in Appendix~\ref{app:ablation} reports the final results.
 
\section{Conclusion}
\label{sec:conclusion}
We introduced \method, which builds the regression target of a one-step flow-map policy by transporting the action gradient of the critic along the behavior flow.
This target recovers the lean adjoint of the underlying optimal control problem exactly when the behavior velocity is affine, and in general its error is bounded by a term that decays inversely with the number of flow-map jumps.
\method requires no best-of-$N$ selection, during training or at test time.
In our experiments, \method outperforms prior one-step methods and is comparable to strong multi-step flow policies on hard tasks with headroom.
A first direction for future work is a control cost budget that accounts for the reliability of the critic.
The learning signal of \method is the action gradient of the critic, and the budget could be reduced automatically where this gradient is less reliable.
A second direction is to deploy \method on real robots and on large vision-language-action policies.
As a one-step policy that needs no selection, \method is well suited to deployment where latency matters.
 
\subsubsection*{Use of Large Language Models}
We used large language models as assistive tools for code development, checking derivations, and polishing the writing. The research ideas are our own, and no experimental result was generated by these models. We take full responsibility for all content.

\bibliography{main}
\bibliographystyle{plainnat}

\clearpage
\appendix
\numberwithin{theorem}{section}
\numberwithin{equation}{section}

\section{Derivations for Section~\ref{sec:method}}
\label{app:proofs}

This appendix contains the derivations behind Section~\ref{sec:method}. Appendix~\ref{app:fixed_point} gives the derivation behind the fixed-point statement of Section~\ref{sec:surrogate}. Appendix~\ref{app:transport} proves Theorem~\ref{thm:transport}: an exact identity for the error of the transport operator gives a bound for arbitrary knots (Proposition~\ref{prop:transport_general}), from which Theorem~\ref{thm:transport} and Corollary~\ref{cor:residual} follow; a numerical illustration closes the section. Appendix~\ref{app:dual} proves Proposition~\ref{prop:dual}.

\textbf{Conventions.} We use the notation of Section~\ref{sec:method} throughout. In Appendices~\ref{app:fixed_point} and~\ref{app:transport} the state $s$ is fixed and omitted, $\nabla Q$ stands for $\nabla_a Q(s, \cdot)$, and $\| \cdot \|$ is the Euclidean norm of a vector and the operator norm of a matrix.

\textbf{Assumptions.} The results of this appendix use the following assumptions.

\begin{assumption}[Regularity]
\label{ass:regularity}
The velocity $b$ and the control $u$ are continuous in $(x, \tau)$ and locally Lipschitz in $x$, $\partial_x b$ exists and is continuous in $(x, \tau)$, and the solutions of $\dot{x}_\tau = b(x_\tau, \tau)$ and of $\dot{x}_\tau = b(x_\tau, \tau) + u(x_\tau, \tau)$ exist on $[0, 1]$ for every initial point.
\end{assumption}

\begin{assumption}[Flow maps]
\label{ass:flow_maps}
$X^\beta$ and $X^\theta$ satisfy the flow-map conditions of Section~\ref{sec:prelim}, that is, they are the flow maps of the velocities $b$ and $b + u$.
\end{assumption}

\begin{assumption}[Bounds on the behavior velocity]
\label{ass:bounds}
For all $x$ and $\tau$, the eigenvalues of the symmetric part of $\partial_x b(x, \tau)$ are at most $\mu$, and $\partial_x b(\cdot, \tau)$ is $L_2$-Lipschitz.
\end{assumption}

\begin{assumption}[Control class]
\label{ass:controls}
The suprema of Section~\ref{sec:algorithm} are taken over a fixed class of controls on which $R$ and $\gC$ are finite.
\end{assumption}

Assumption~\ref{ass:regularity} is the standard regularity under which the two flows and their Jacobians are well defined. It is the regularity that Theorem~\ref{thm:transport} refers to, and Assumptions~\ref{ass:flow_maps} and~\ref{ass:bounds} are the hypotheses that the theorem states. Appendix~\ref{app:fixed_point} assumes in addition that $u$ is continuously differentiable in $x$. The constant $\mu$ may be negative, and $L_2 = 0$ if $b$ is affine in $x$. Assumption~\ref{ass:bounds} is needed for the bounds only: the identities of Proposition~\ref{prop:transport_general}~(i) hold without it. Assumption~\ref{ass:controls} is used in Appendix~\ref{app:dual} only.

\textbf{Propagator.} Along the policy trajectory $x_\tau = X^\theta_{t,\tau}(x_t)$ through a point $x_t$, let $\Phi(\tau, \sigma)$, $t \le \sigma \le \tau \le 1$, be the solution of
\begin{equation}
  \partial_\tau \Phi(\tau, \sigma) = \partial_x b(x_\tau, \tau)\, \Phi(\tau, \sigma), \qquad \Phi(\sigma, \sigma) = I .
  \label{eq:propagator}
\end{equation}
It satisfies $\Phi(1, \tau)\, \Phi(\tau, \sigma) = \Phi(1, \sigma)$ and therefore $\partial_\tau \Phi(1, \tau) = - \Phi(1, \tau)\, \partial_x b(x_\tau, \tau)$. The solution of \eqref{eq:lean} is $\tilde{p}_\tau = \lambda\, \Phi(1, \tau)^\top \nabla Q(x_1)$, so that $T^\star_t = \Phi(1, t)^\top$.

\subsection{Lean adjoint and value gradient}
\label{app:fixed_point}

Lean adjoint matching and value-gradient matching have the same fixed points \citep{liu2026valuegradientguidanceflow}. To keep the paper self-contained, we give a short derivation in our notation.

\textbf{Value gradient of a control.} Let $u$ be continuously differentiable in $x$, and let $V^u(x_t, t) = \lambda\, Q(x_1) - \frac{1}{2} \int_t^1 \| u(x_\tau, \tau) \|^2 \, d\tau$ be the value of the control $u$ on its trajectory through $x_t$. Write $\Psi_\tau = \partial_x X^\theta_{t,\tau}(x_t)$, $u_\tau = u(x_\tau, \tau)$ and $U_\tau = \partial_x u(x_\tau, \tau)$. Differentiating the Lagrangian condition of $X^\theta$ in $x$ gives $\partial_\tau \Psi_\tau = \big( \partial_x b(x_\tau, \tau) + U_\tau \big) \Psi_\tau$ with $\Psi_t = I$, and by variation of constants
\begin{equation}
  \Psi_1 = \Phi(1, t) + \int_t^1 \Phi(1, \tau)\, U_\tau \Psi_\tau \, d\tau .
  \label{eq:variation_of_constants}
\end{equation}
Differentiating $V^u$ along the trajectory, and then inserting \eqref{eq:variation_of_constants} and $\lambda\, \Phi(1, \tau)^\top \nabla Q(x_1) = \tilde{p}_\tau$,
\begin{equation}
  \nabla_x V^u(x_t, t) = \lambda\, \Psi_1^\top \nabla Q(x_1) - \int_t^1 \Psi_\tau^\top U_\tau^\top u_\tau \, d\tau = \tilde{p}_t + \int_t^1 \Psi_\tau^\top U_\tau^\top \big( \tilde{p}_\tau - u_\tau \big) \, d\tau .
  \label{eq:value_vs_lean}
\end{equation}
The lean adjoint omits the integral, the only term that contains Jacobians of the policy.

\textbf{Equivalence of the fixed points.} If $u_\tau = \tilde{p}_\tau$ on $[t, 1]$, the integral of \eqref{eq:value_vs_lean} vanishes at every point of the trajectory, so that $\nabla_x V^u(x_\tau, \tau) = \tilde{p}_\tau = u_\tau$. Conversely, if $u_\tau = \nabla_x V^u(x_\tau, \tau)$ on $[t, 1]$, then \eqref{eq:value_vs_lean}, applied at $x_\tau$, shows that $e_\tau = \tilde{p}_\tau - u_\tau$ satisfies
\begin{equation}
  e_\tau = - \int_\tau^1 \big[ \partial_x X^\theta_{\tau,\sigma}(x_\tau) \big]^\top U_\sigma^\top e_\sigma \, d\sigma , \qquad \text{hence} \qquad \| e_\tau \| \le C \int_\tau^1 \| e_\sigma \| \, d\sigma
\end{equation}
for a constant $C$, and $e = 0$ by Gr\"onwall's inequality. A control therefore equals its lean adjoint along its trajectories if and only if it equals the gradient of its own value function.

\textbf{Stationary points.} These controls are the stationary points of the control problem in \eqref{eq:oc}, whose objective is $\E_{x_0} [ V^u(x_0, 0) ]$. For a perturbation $u + \epsilon w$ of the control, the trajectory changes to first order by $\frac{d}{d\epsilon}\, x_\sigma = \int_0^\sigma \partial_x X^\theta_{\tau,\sigma}(x_\tau)\, w(x_\tau, \tau) \, d\tau$, and with the first equality of \eqref{eq:value_vs_lean},
\begin{equation}
  \frac{d}{d\epsilon}\, \E_{x_0} \big[ V^{u + \epsilon w}(x_0, 0) \big] \Big|_{\epsilon = 0} = \E_{x_0} \int_0^1 w(x_\tau, \tau)^\top \big( \nabla_x V^u(x_\tau, \tau) - u_\tau \big) \, d\tau .
  \label{eq:first_variation}
\end{equation}
The first variation vanishes for every $w$ if and only if $u = \nabla_x V^u$ along the trajectories of $u$.

\subsection{Proof of Theorem~\ref{thm:transport}}
\label{app:transport}

We restate the theorem of Section~\ref{sec:surrogate}.

\begin{restatedthm}
Under the regularity conditions of Appendix~\ref{app:proofs}, assume that $X^\beta$ and $X^\theta$ satisfy the flow-map conditions of Section~\ref{sec:prelim}, that is, they are the flow maps of the velocities $b$ and $b + u$. Assume further that, for all $x$ and $\tau$, the eigenvalues of the symmetric part of $\partial_x b(x, \tau)$ are at most $\mu$ and $\partial_x b(\cdot, \tau)$ is $L_2$-Lipschitz. Then, for every $t$ and $x_t$ and for $K$ evenly spaced knots with spacing $h = (1 - t) / K$,
\begin{equation}
  \big\| T_t - T^\star_t \big\| \;\le\; \frac{L_2\, (1 - t)^2}{2 K}\; e^{\mu (1 - t) + \max(\mu, 0)\, h}\, \sup_{\tau \in [t, 1]} \big\| u(x_\tau, \tau) \big\| ,
  \tag{\ref{eq:transport_bound}}
\end{equation}
where $x_\tau = X^\theta_{t,\tau}(x_t)$ is the trajectory of the policy through $x_t$ and the norm on the left is the operator norm. In particular, $T_t = T^\star_t$ for every $K$ if $b$ is affine in $x$.
\end{restatedthm}

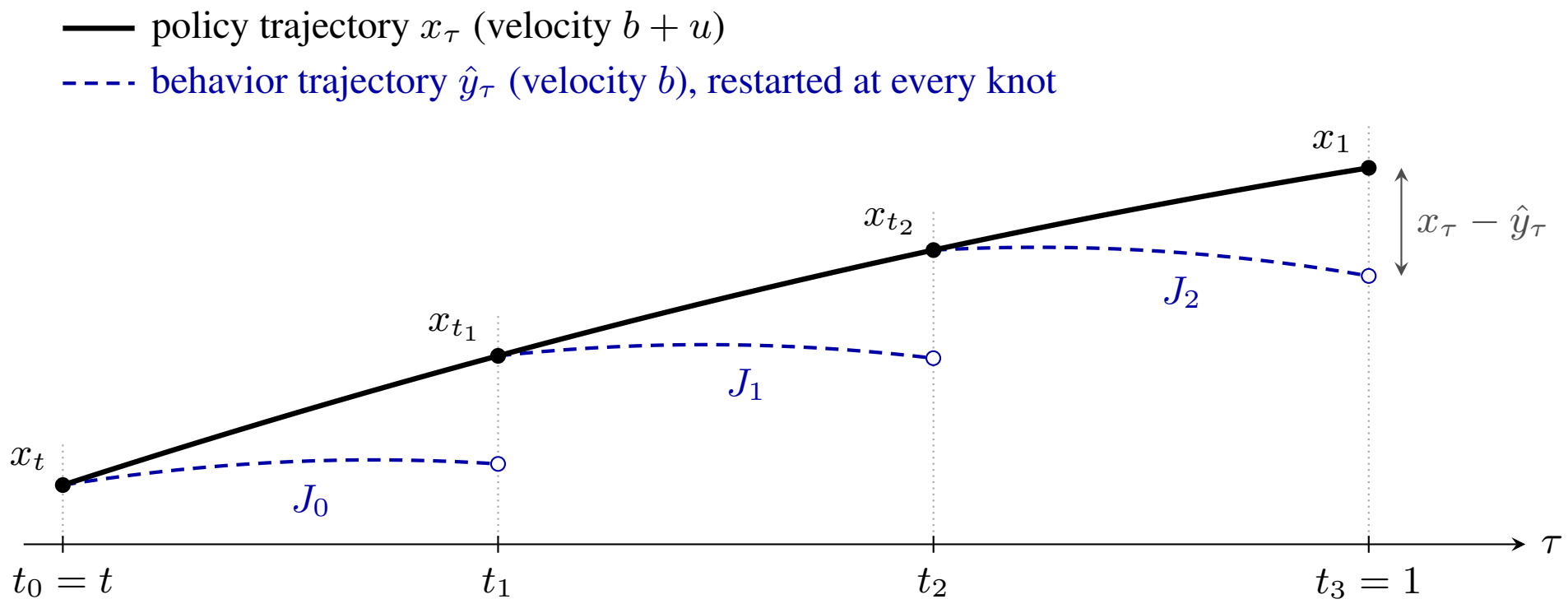
\begin{figure}[h]
\begin{center}
\begin{tikzpicture}[x=10.5cm, y=3.4cm, >=stealth, font=\small,
    declare function={pol(\x) = 0.14 + 1.0*\x - 0.25*\x*\x; beh(\x,\a) = pol(\x) - 0.5*(\x-\a) - 0.8*(\x-\a)*(\x-\a);}]
  \draw[->] (-0.03, 0) -- (1.12, 0) node[right] {$\tau$};
  \foreach \a/\l in {0/{t_0 = t}, 0.3333/{t_1}, 0.6667/{t_2}, 1/{t_3 = 1}} {
    \draw[densely dotted, black!35] (\a, 0) -- (\a, {pol(\a) + 0.1});
    \draw (\a, 0.02) -- (\a, -0.02) node[below] {$\l$};
  }
  \foreach \a/\k in {0/0, 0.3333/1, 0.6667/2} {
    \draw[thick, densely dashed, blue!65!black] plot[domain=\a:\a+0.3333, samples=20] (\x, {beh(\x,\a)});
    \draw[blue!65!black, fill=white] ({\a+0.3333}, {beh(\a+0.3333,\a)}) circle (1.6pt);
    \node[blue!65!black, below] at ({\a+0.19}, {beh(\a+0.19,\a) - 0.02}) {$J_{\k}$};
  }
  \draw[very thick] plot[domain=0:1, samples=40] (\x, {pol(\x)});
  \foreach \a/\l in {0/{x_t}, 0.3333/{x_{t_1}}, 0.6667/{x_{t_2}}, 1/{x_1}} {
    \fill (\a, {pol(\a)}) circle (1.8pt);
    \node[above left] at (\a, {pol(\a)}) {$\l$};
  }
  \draw[<->, black!70] (1.025, {beh(1,0.6667)}) -- (1.025, {pol(1)}) node[midway, right] {$x_\tau - \hat{y}_\tau$};
  \draw[very thick] (0.0, 1.22) -- (0.055, 1.22) node[right, black] {policy trajectory $x_\tau$ (velocity $b + u$)};
  \draw[thick, densely dashed, blue!65!black] (0.0, 1.09) -- (0.055, 1.09) node[right] {behavior trajectory $\hat{y}_\tau$ (velocity $b$), restarted at every knot};
\end{tikzpicture}
\end{center}
\caption{The two trajectories in the proof of Theorem~\ref{thm:transport}, for $K = 3$ jumps; $J_k$ is the Jacobian of the $k$-th behavior jump, \eqref{eq:transport}.}
\label{fig:restart}
\end{figure}

\textbf{Idea of the proof.} Both operators are propagators of the same linear equation, $M' = \partial_x b\, M$, along two different trajectories (Figure~\ref{fig:restart}). The lean adjoint uses the policy trajectory $x_\tau$, whereas the Jacobian of the behavior jump $X^\beta_{t_k, t_{k+1}}$ at $x_{t_k}$ uses the behavior trajectory that starts at $x_{t_k}$. The error of $T_t$ is therefore driven by the distance between the policy trajectory and a behavior trajectory that is restarted at every knot, and this distance is the control accumulated since the last knot.

\textbf{Setup.} The knots $t_k$ of \eqref{eq:transport} need not be evenly spaced here; let $h_k = t_{k+1} - t_k$ and $h = \max_k h_k$. For $\tau \in [t_k, t_{k+1})$, let $\hat{y}_\tau = X^\beta_{t_k,\tau}(x_{t_k})$ be the behavior trajectory restarted at the last knot, and for $\tau \in [t_k, t_{k+1}]$ let
\begin{equation}
  \hat{\Phi}(\tau) = \partial_x X^\beta_{t_k,\tau}(x_{t_k})\, J_{k-1} \cdots J_0 ,
  \label{eq:surrogate_propagator}
\end{equation}
with $J_k$ as in \eqref{eq:transport}. The function $\hat{\Phi}$ is continuous, $\hat{\Phi}(t) = I$ and $\hat{\Phi}(1) = T_t^\top$, and differentiating the Lagrangian condition of $X^\beta$ in $x$ gives $\partial_\tau \hat{\Phi}(\tau) = \partial_x b(\hat{y}_\tau, \tau)\, \hat{\Phi}(\tau)$ between two knots. Let $\phi_\mu(z) = (e^{\mu z} - 1) / \mu$ for $\mu \ne 0$ and $\phi_0(z) = z$.

\begin{lemma}[Growth bound]
\label{lem:growth}
Let $M(\tau)$ be a continuous and piecewise continuously differentiable matrix function with $M' = A M$, where the symmetric part of $A(\tau)$ has eigenvalues at most $\mu$. Then $\| M(\tau)\, \xi \| \le e^{\mu (\tau - \sigma)}\, \| M(\sigma)\, \xi \|$ for every vector $\xi$ and all $\sigma \le \tau$. In particular, under Assumptions~\ref{ass:regularity}--\ref{ass:bounds},
\begin{equation}
  \| \Phi(1, \tau) \| \le e^{\mu (1 - \tau)} , \qquad \| \hat{\Phi}(\tau) \| \le e^{\mu (\tau - t)} , \qquad \big\| \partial_x X^\beta_{\sigma,\tau}(x_\sigma) \big\| \le e^{\mu (\tau - \sigma)} .
  \label{eq:growth}
\end{equation}
\end{lemma}

\begin{proof}
Wherever $M$ is differentiable, $\frac{d}{d\tau} \| M \xi \|^2 = 2\, (M \xi)^\top A\, (M \xi) \le 2 \mu\, \| M \xi \|^2$, which gives the first claim. The three matrices of \eqref{eq:growth} solve such an equation with $A = \partial_x b$ evaluated along a trajectory, and they are the identity at $\tau$, at $t$ and at $\sigma$, respectively.
\end{proof}

\begin{proposition}[Arbitrary knots]
\label{prop:transport_general}
Let Assumptions~\ref{ass:regularity} and~\ref{ass:flow_maps} hold. For every $t$, every $x_t$ and every choice of the knots:
\begin{enumerate}
\item[(i)] The error of the transport operator and the distance between the two trajectories are
\begin{align}
  T_t^\top - \Phi(1, t) &= \int_t^1 \Phi(1, \tau) \big[ \partial_x b(\hat{y}_\tau, \tau) - \partial_x b(x_\tau, \tau) \big] \hat{\Phi}(\tau) \, d\tau ,
  \label{eq:error_identity} \\
  x_\tau - \hat{y}_\tau &= \int_{t_k}^{\tau} \partial_x X^\beta_{\sigma,\tau}(x_\sigma)\, u(x_\sigma, \sigma) \, d\sigma \qquad \text{for } \tau \in [t_k, t_{k+1}) .
  \label{eq:path_deviation}
\end{align}
\item[(ii)] Under Assumption~\ref{ass:bounds},
\begin{equation}
  \big\| T_t - T^\star_t \big\| \;\le\; L_2\, e^{\mu (1 - t)} \sum_{k=0}^{K-1} \int_{t_k}^{t_{k+1}} \phi_\mu(t_{k+1} - \tau)\, \big\| u(x_\tau, \tau) \big\| \, d\tau .
  \label{eq:transport_bound_general}
\end{equation}
\item[(iii)] $T_t = T^\star_t$ if $b$ is affine in $x$, or if the control vanishes along the trajectory.
\end{enumerate}
\end{proposition}

\begin{proof}
\emph{Step 1 (error identity).} The function $G(\tau) = \Phi(1, \tau)\, \hat{\Phi}(\tau)$ is continuous on $[t, 1]$ with $G(t) = \Phi(1, t)$ and $G(1) = T_t^\top$. Between two knots, $\partial_\tau \Phi(1, \tau) = - \Phi(1, \tau)\, \partial_x b(x_\tau, \tau)$ and $\partial_\tau \hat{\Phi}(\tau) = \partial_x b(\hat{y}_\tau, \tau)\, \hat{\Phi}(\tau)$, so the product rule gives
\begin{equation}
  G'(\tau) = \Phi(1, \tau) \big[ \partial_x b(\hat{y}_\tau, \tau) - \partial_x b(x_\tau, \tau) \big] \hat{\Phi}(\tau) .
\end{equation}
Integrating between consecutive knots and summing gives \eqref{eq:error_identity}.

\emph{Step 2 (distance between the trajectories).} Fix $\tau \in [t_k, t_{k+1}]$ and let $\zeta(\sigma) = X^\beta_{\sigma,\tau}(x_\sigma)$ for $\sigma \in [t_k, \tau]$, so that $\zeta(\tau) = x_\tau$ and $\zeta(t_k) = X^\beta_{t_k,\tau}(x_{t_k})$. By the chain rule, the Lagrangian condition of $X^\theta$ and the Eulerian condition of $X^\beta$,
\begin{equation}
  \zeta'(\sigma) = \partial_\sigma X^\beta_{\sigma,\tau}(x_\sigma) + \partial_x X^\beta_{\sigma,\tau}(x_\sigma) \big[ b(x_\sigma, \sigma) + u(x_\sigma, \sigma) \big] = \partial_x X^\beta_{\sigma,\tau}(x_\sigma)\, u(x_\sigma, \sigma) ,
\end{equation}
and integrating over $[t_k, \tau]$ gives \eqref{eq:path_deviation}.

\emph{Step 3 (bound).} In \eqref{eq:error_identity}, Lemma~\ref{lem:growth} gives $\| \Phi(1, \tau) \|\, \| \hat{\Phi}(\tau) \| \le e^{\mu (1 - \tau)}\, e^{\mu (\tau - t)} = e^{\mu (1 - t)}$, and Assumption~\ref{ass:bounds} gives $\| \partial_x b(\hat{y}_\tau, \tau) - \partial_x b(x_\tau, \tau) \| \le L_2\, \| x_\tau - \hat{y}_\tau \|$. In \eqref{eq:path_deviation}, Lemma~\ref{lem:growth} bounds the Jacobian of the behavior flow map. Hence
\begin{align}
  \big\| T_t^\top - \Phi(1, t) \big\| &\le L_2\, e^{\mu (1 - t)} \int_t^1 \| x_\tau - \hat{y}_\tau \| \, d\tau , \\
  \| x_\tau - \hat{y}_\tau \| &\le \int_{t_k}^{\tau} e^{\mu (\tau - \sigma)}\, \| u(x_\sigma, \sigma) \| \, d\sigma \qquad \text{for } \tau \in [t_k, t_{k+1}) .
\end{align}
Exchanging the order of integration between two knots and using $\int_\sigma^{t_{k+1}} e^{\mu (\tau - \sigma)} \, d\tau = \phi_\mu(t_{k+1} - \sigma)$,
\begin{equation}
  \int_{t_k}^{t_{k+1}} \! \int_{t_k}^{\tau} e^{\mu (\tau - \sigma)} \| u(x_\sigma, \sigma) \| \, d\sigma \, d\tau = \int_{t_k}^{t_{k+1}} \phi_\mu(t_{k+1} - \sigma)\, \| u(x_\sigma, \sigma) \| \, d\sigma ,
\end{equation}
which proves (ii), since the operator norm is invariant under transposition.

\emph{Step 4 (exact cases).} If $b$ is affine in $x$, then $\partial_x b$ does not depend on $x$ and the right-hand side of \eqref{eq:error_identity} vanishes. If the control vanishes along the trajectory, \eqref{eq:path_deviation} gives $\hat{y}_\tau = x_\tau$, and the right-hand side of \eqref{eq:error_identity} vanishes again.
\end{proof}

\begin{proof}[Proof of Theorem~\ref{thm:transport}]
We first bound $\int_0^{h_k} \phi_\mu(z) \, dz$, which equals $(e^{\mu h_k} - 1 - \mu h_k) / \mu^2$ for $\mu \ne 0$ and $h_k^2 / 2$ for $\mu = 0$. For $z > 0$,
\begin{equation}
  e^{z} - 1 - z = \sum_{n \ge 0} \frac{z^{n+2}}{(n+2)!} \le \frac{z^2}{2} \sum_{n \ge 0} \frac{z^n}{n!} = \frac{z^2}{2}\, e^{z} \qquad \text{and} \qquad e^{-z} - 1 + z \le \frac{z^2}{2} ,
\end{equation}
where the second inequality holds because the difference of its two sides vanishes at $z = 0$ and has the derivative $z - 1 + e^{-z} \ge 0$. With $z = \mu h_k$ for $\mu > 0$ and $z = - \mu h_k$ for $\mu < 0$,
\begin{equation}
  \int_0^{h_k} \phi_\mu(z) \, dz \le \frac{h_k^2}{2}\, e^{\max(\mu, 0)\, h_k} .
\end{equation}
Bounding $\| u(x_\tau, \tau) \|$ by its supremum in \eqref{eq:transport_bound_general} and using $\sum_k h_k^2 = (1 - t)^2 / K$ for evenly spaced knots gives \eqref{eq:transport_bound}. The exact case is Proposition~\ref{prop:transport_general}~(iii).
\end{proof}

\begin{corollary}[Residual of the stationarity condition]
\label{cor:residual}
Under Assumptions~\ref{ass:regularity}--\ref{ass:bounds}, suppose that the policy is a fixed point of \eqref{eq:guide} along the trajectory, that is, $u(x_\tau, \tau) = \lambda\, T_\tau \nabla Q(x_1)$ for all $\tau \in [t, 1]$, where $T_\tau$ is built on $K$ evenly spaced knots of $[\tau, 1]$. Then
\begin{equation}
  \big\| u(x_t, t) - \tilde{p}_t \big\| \;\le\; \frac{L_2\, (1 - t)^2}{2 K}\; e^{(\mu + \max(\mu, 0)) (1 - t) + \max(\mu, 0)\, h}\; \lambda^2\, \big\| \nabla Q(x_1) \big\|^2 .
  \label{eq:residual}
\end{equation}
\end{corollary}

\begin{proof}
The matrix $T_\tau^\top$ is a product of Jacobians of behavior jumps that cover $[\tau, 1]$, so $\| T_\tau \| \le e^{\mu (1 - \tau)} \le e^{\max(\mu, 0) (1 - t)}$ by Lemma~\ref{lem:growth}, and $\| u(x_\tau, \tau) \| \le \lambda\, e^{\max(\mu, 0) (1 - t)}\, \| \nabla Q(x_1) \|$. Since $u(x_t, t) - \tilde{p}_t = \lambda\, (T_t - T^\star_t)\, \nabla Q(x_1)$, the claim follows from Theorem~\ref{thm:transport}.
\end{proof}

The residual is thus of second order in $\lambda\, \| \nabla Q \|$, whereas the guidance term is of first order.

\begin{remark}[Constants]
\label{rem:constants}
The proof uses the bound $\mu$ only at the points $X^\beta_{\sigma,\tau}(x_\sigma)$ with $t_k \le \sigma \le \tau \le t_{k+1}$, and the Lipschitz bound only for the pairs $(x_\tau, \hat{y}_\tau)$. The global constants of Theorem~\ref{thm:transport} can therefore be replaced by the constants of $b$ along these trajectories.
\end{remark}

\begin{remark}[Control cost]
The bound can also be stated with the control cost on the trajectory. Since $\phi_\mu(z) \le z\, e^{\max(\mu, 0)\, z}$ for $z \ge 0$, evenly spaced knots give $\sum_k \int_0^{h} \phi_\mu(z)^2 \, dz \le \frac{1}{3} (1 - t)\, h^2\, e^{2 \max(\mu, 0)\, h}$, and the Cauchy--Schwarz inequality in \eqref{eq:transport_bound_general} gives
\begin{equation}
  \big\| T_t - T^\star_t \big\| \;\le\; \frac{L_2\, (1 - t)^{3/2}}{K}\; e^{\mu (1 - t) + \max(\mu, 0)\, h}\, \Big( \frac{1}{3} \int_t^1 \big\| u(x_\tau, \tau) \big\|^2 \, d\tau \Big)^{1/2} .
\end{equation}
\end{remark}

\begin{remark}[Sharpness]
\label{rem:sharpness}
The bound of \eqref{eq:transport_bound_general} cannot be improved in general, as the following example shows. In dimension one, let $b(x, \tau) = m x + \frac{q}{2} x^2$ with $q > 0$, let the control $u = c > 0$ be constant and let $x_t = 0$. The velocity $b + c$ of the policy does not depend on $\tau$ and equals $c > 0$ at $x_t = 0$, so the policy trajectory is increasing and remains in $[0, x_1]$, with $x_1 = O(c)$ as $c \to 0$. Since $b(0, \tau) = 0$, a comparison of scalar ODEs gives $0 \le X^\beta_{\sigma,\tau}(x_\sigma) \le x_\tau$ for the behavior trajectories that start on it. All trajectories used in the proof therefore remain in $[0, x_1]$, where $L_2 = q$ and $\mu = m + q\, x_1$ (Remark~\ref{rem:constants}). Since $\partial_x b(x, \tau) = m + q\, x$,
\begin{equation}
  \Phi(1, t) = \exp \Big( \int_t^1 ( m + q\, x_\tau ) \, d\tau \Big) , \qquad T_t = \exp \Big( \int_t^1 ( m + q\, \hat{y}_\tau ) \, d\tau \Big) ,
\end{equation}
so that
\begin{equation}
  \big| T_t - T^\star_t \big| = \Phi(1, t)\, \Big[ 1 - \exp \Big( - q \int_t^1 ( x_\tau - \hat{y}_\tau ) \, d\tau \Big) \Big] .
\end{equation}
Here $\Phi(1, t) = e^{m (1 - t)} (1 + O(c))$, and \eqref{eq:path_deviation} gives $x_\tau - \hat{y}_\tau = c\, \phi_m(\tau - t_k)\, (1 + O(c))$ for $\tau \in [t_k, t_{k+1})$. With $\psi_\mu(z) = \int_0^z \phi_\mu(z') \, dz'$, the left-hand side of \eqref{eq:transport_bound_general} is therefore $q\, c\, e^{m (1 - t)} \sum_k \psi_m(h_k)\, (1 + O(c))$, and its right-hand side is $q\, c\, e^{\mu (1 - t)} \sum_k \psi_\mu(h_k)$ with $\mu = m + O(c)$. Their ratio tends to one as $c \to 0$, for every $K$ and every placement of the knots.
\end{remark}

\begin{remark}[Euler steps]
\label{rem:euler}
An Euler step of the behavior velocity is a first-order approximation of a behavior jump, and its Jacobian $J^{\mathrm{E}}_k = I + h_k\, \partial_x b(x_{t_k}, t_k)$ approximates the Jacobian $J_k$ of \eqref{eq:transport}. The approximation saves no computation. In terms of the average velocity of Section~\ref{sec:behavior}, $J_k = I + h_k\, \partial_x v_\beta(x_{t_k}, t_k, t_{k+1})$ and $J^{\mathrm{E}}_k = I + h_k\, \partial_x v_\beta(x_{t_k}, t_k, t_k)$, so both take one vector-Jacobian product of the same network. Let $b$ be continuously differentiable in $(x, \tau)$. Along a behavior trajectory $y_\tau$, where $\dot{y}_\tau = b(y_\tau, \tau)$, the chain rule gives the acceleration $\ddot{y}_\tau = \dot{b}(y_\tau, \tau)$ with $\dot{b}(x, \tau) = \partial_\tau b(x, \tau) + \partial_x b(x, \tau)\, b(x, \tau)$. If $\dot{b}$ is continuously differentiable in $x$, Taylor's formula with integral remainder for $\tau \mapsto X^\beta_{t_k,\tau}(x)$, differentiated in $x$ at $x_{t_k}$, gives
\begin{equation}
  J_k - J^{\mathrm{E}}_k = \int_{t_k}^{t_{k+1}} (t_{k+1} - \tau)\, \partial_x \dot{b}(\hat{y}_\tau, \tau)\, \partial_x X^\beta_{t_k,\tau}(x_{t_k}) \, d\tau .
  \label{eq:euler_identity}
\end{equation}
Euler steps thus add an error of order $h_k^2$ to each factor of the transport operator, which, unlike the error of Theorem~\ref{thm:transport}, does not vanish with the control. It vanishes if $\dot{b} = 0$, that is, if the behavior trajectories are straight lines traversed at constant velocity. The exact cases of Proposition~\ref{prop:transport_general}~(iii) do not carry over to Euler steps. In dimension one with $b(x, \tau) = m x$, the behavior jumps give $J_k = e^{m h_k}$ and $T_t = T^\star_t = e^{m (1 - t)}$ for every control, whereas $J^{\mathrm{E}}_k = 1 + m h_k$.
\end{remark}

\begin{figure}[h]
\begin{center}
\includegraphics[width=0.8\linewidth]{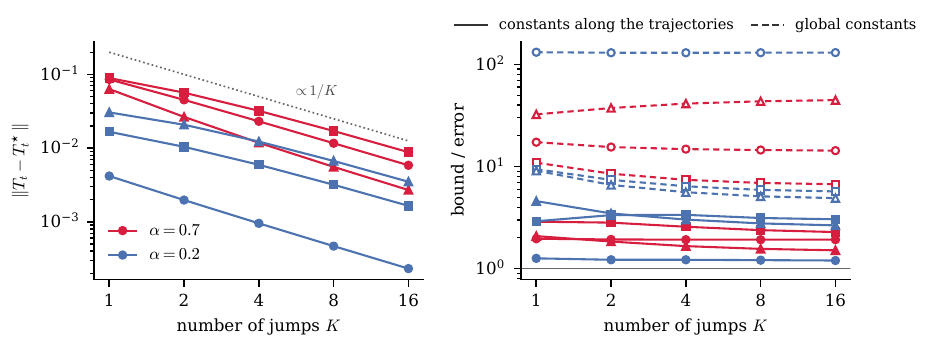}
\end{center}
\caption{Error of the transport operator on the three-dimensional example; colors are the control amplitudes $\alpha$ and markers the starting points $x_t$. Left: $\| T_t - T^\star_t \|$ against the number of jumps $K$. Right: the bound of \eqref{eq:transport_bound_general} divided by the error.}
\label{fig:transport_error}
\end{figure}

\textbf{Numerical illustration.} Figure~\ref{fig:transport_error} evaluates the two sides of \eqref{eq:transport_bound_general} on a three-dimensional example with $b(x, \tau) = - \frac{1}{2} x + A \tanh(W x) + \tau\, c$ and the feedback control $u(x, \tau) = \alpha \tanh(V x) + \frac{\alpha}{0.7}\, \tau\, e$, where the matrices $A$, $W$, $V$ and the vectors $c$, $e$ are drawn once from a Gaussian distribution, $t = 0.2$, the knots are evenly spaced, and all trajectories and propagators are integrated with a fourth-order Runge--Kutta scheme. Over two control amplitudes $\alpha$, three starting points $x_t$ and $K \in \{1, 2, 4, 8, 16\}$, the identity of \eqref{eq:error_identity} holds up to $5 \times 10^{-15}$ and the bound is never violated. The error decreases at the rate $1 / K$: from $K = 8$ to $K = 16$ it shrinks by a factor between $1.91$ and $2.06$. The bound exceeds the error by a factor between $4.9$ and $132$ with the global constants of this example, $\mu = 0.39$ and $L_2 = 3.72$, and by a factor between $1.2$ and $4.6$ with the constants along the trajectories (Remark~\ref{rem:constants}).

\subsection{Proof of Proposition~\ref{prop:dual}}
\label{app:dual}

We use the notation of Section~\ref{sec:algorithm} and Assumption~\ref{ass:controls}, and restate the proposition of Section~\ref{sec:algorithm}.

\begin{restatedprop}
\textit{(i)} For every $\rho > 0$ and every control $u$, $\ell(u, \rho) = \rho\, \delta + \rho\, \gJ_{1/\rho}(u)$. Hence $u$ maximizes $\ell(\cdot, \rho)$ if and only if it maximizes $\gJ_\lambda$ with $\lambda = 1 / \rho$, and such a control solves \eqref{eq:budget} if its control cost equals $\delta$.
\textit{(ii)} The dual function $D$ is convex, and $D(\rho)$ is an upper bound on the value of \eqref{eq:budget} for every $\rho \ge 0$. If $u_\rho$ maximizes $\ell(\cdot, \rho)$, then $\delta - \gC(u_\rho)$ is a subgradient of $D$ at $\rho$.
\end{restatedprop}

\begin{proof}
\textit{(i)} For $\rho > 0$,
\begin{equation}
  \ell(u, \rho) = \rho\, \delta + \rho\, \Big( \frac{1}{\rho}\, R(u) - \gC(u) \Big) = \rho\, \delta + \rho\, \gJ_{1/\rho}(u) .
\end{equation}
The first term does not depend on $u$ and $\rho$ is positive, so $\ell(\cdot, \rho)$ and $\gJ_{1/\rho}$ have the same maximizers. Let $u_\rho$ be one of them and suppose that $\gC(u_\rho) = \delta$. Then $u_\rho$ satisfies the constraint of \eqref{eq:budget} and $\ell(u_\rho, \rho) = R(u_\rho)$. For every control $u$ with $\gC(u) \le \delta$,
\begin{equation}
  R(u) \;\le\; R(u) - \rho\, \big( \gC(u) - \delta \big) = \ell(u, \rho) \;\le\; \ell(u_\rho, \rho) = R(u_\rho) ,
\end{equation}
so $u_\rho$ solves \eqref{eq:budget}.

\textit{(ii)} For every control $u$, the function $\rho \mapsto \ell(u, \rho)$ is affine, so $D$ is convex as a supremum of affine functions. For every $\rho \ge 0$ and every control $u$ with $\gC(u) \le \delta$, $R(u) \le \ell(u, \rho) \le D(\rho)$, so $D(\rho)$ is an upper bound on the value of \eqref{eq:budget}. If $u_\rho$ maximizes $\ell(\cdot, \rho)$, then for every $\rho' \ge 0$,
\begin{equation}
  D(\rho') \;\ge\; \ell(u_\rho, \rho') = \ell(u_\rho, \rho) + (\rho' - \rho) \big( \delta - \gC(u_\rho) \big) = D(\rho) + (\rho' - \rho) \big( \delta - \gC(u_\rho) \big) ,
\end{equation}
which is the definition of a subgradient.
\end{proof}

\begin{remark}[Monotonicity]
The control cost of a maximizer of $\gJ_\lambda$ does not decrease with $\lambda$, so a larger budget corresponds to a larger coefficient. If $u$ maximizes $\gJ_\lambda$ and $u'$ maximizes $\gJ_{\lambda'}$ with $\lambda < \lambda'$, then
\begin{equation}
  \lambda\, R(u) - \gC(u) \;\ge\; \lambda\, R(u') - \gC(u') , \qquad \lambda'\, R(u') - \gC(u') \;\ge\; \lambda'\, R(u) - \gC(u) .
\end{equation}
Adding the two inequalities gives $(\lambda' - \lambda) \big( R(u') - R(u) \big) \ge 0$, hence $R(u') \ge R(u)$, and the first inequality then gives $\gC(u') - \gC(u) \ge \lambda\, \big( R(u') - R(u) \big) \ge 0$.
\end{remark}

\begin{remark}[Scope]
The upper bound of \textit{(ii)} need not be attained: the problem is not concave in $u$ in general, and a budget $\delta$ may then not be the control cost of a maximizer of $\gJ_\lambda$ for any $\lambda$. In terms of the multiplier, the update of \eqref{eq:dual} reads $\rho \leftarrow \rho\, \exp \big( - \frac{\eta}{\delta}\, ( \delta - \hat{\gC} ) \big)$, which to first order in $\eta$ is the subgradient step $\rho \leftarrow \rho - \frac{\eta \rho}{\delta}\, ( \delta - \hat{\gC} )$. Proposition~\ref{prop:dual} thus gives the direction of the update, not its convergence.
\end{remark}
 \FloatBarrier
\section{Benchmark and Experiment Details}
\label{app:benchmark}

\begin{figure}[h]
  \centering
  \setlength{\tabcolsep}{2pt}
  \begin{tabular}{ccc}
    \includegraphics[width=0.32\textwidth]{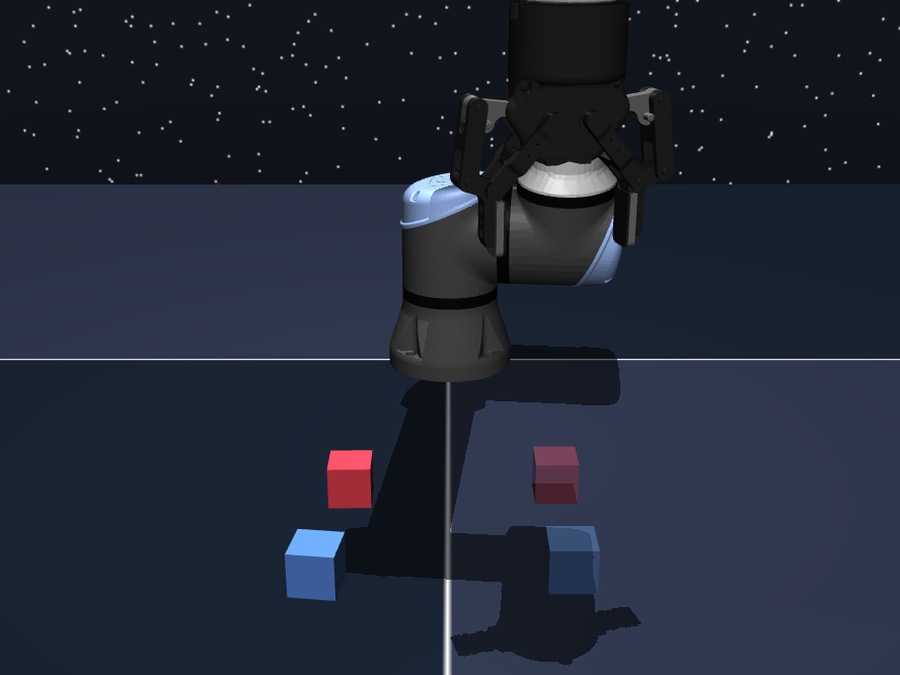} &
    \includegraphics[width=0.32\textwidth]{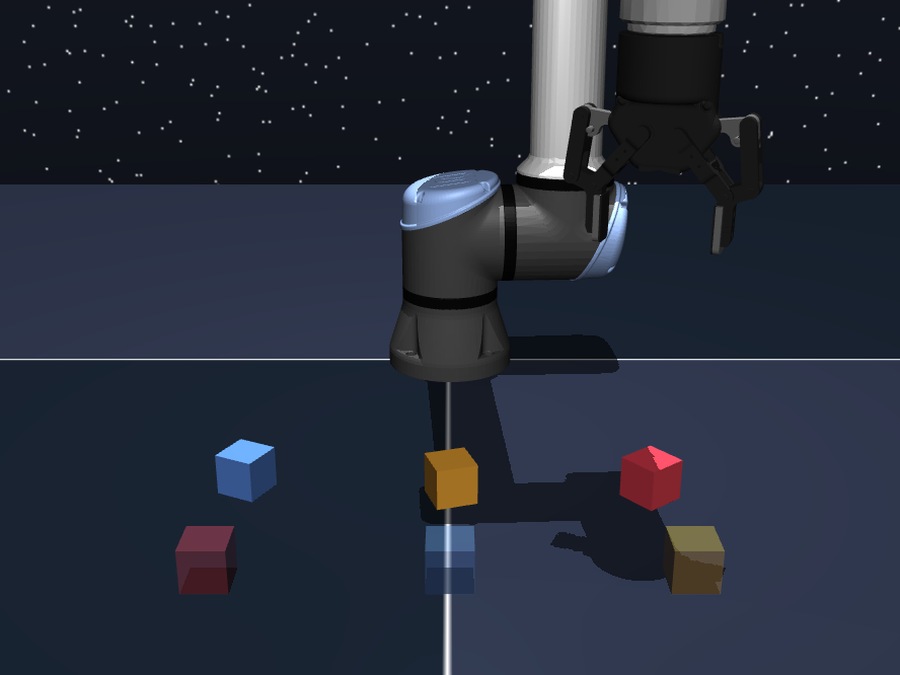} &
    \includegraphics[width=0.32\textwidth]{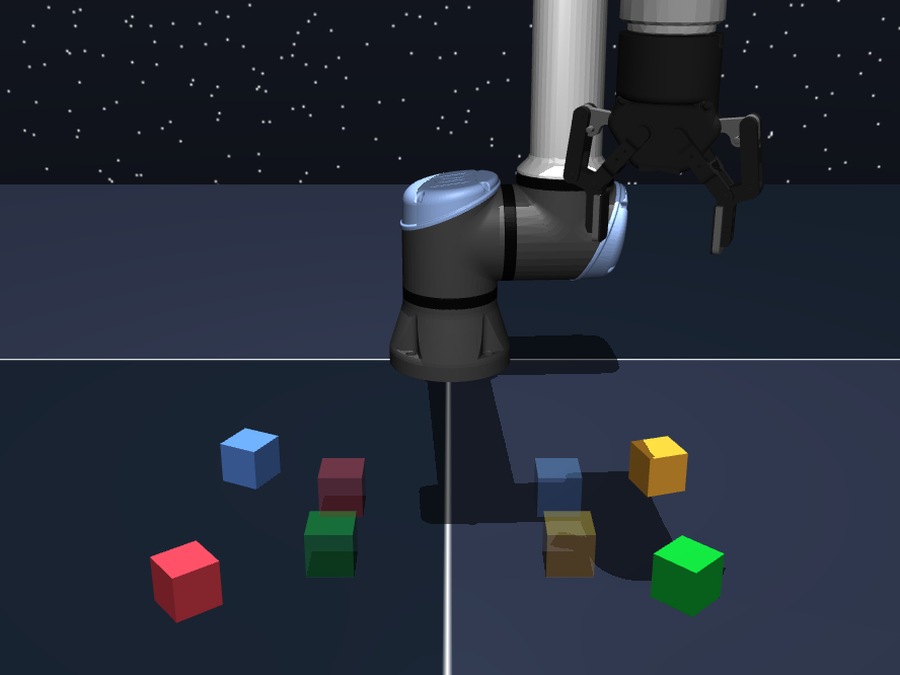} \\
    \texttt{cube-double} & \texttt{cube-triple} & \texttt{cube-quadruple} \\[4pt]
    \includegraphics[width=0.32\textwidth]{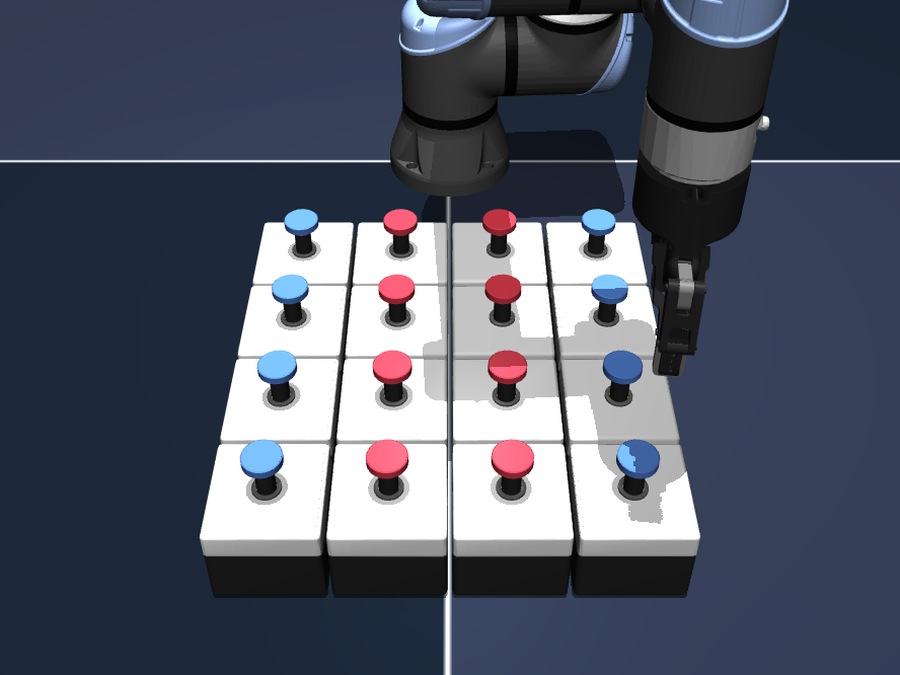} &
    \includegraphics[width=0.32\textwidth]{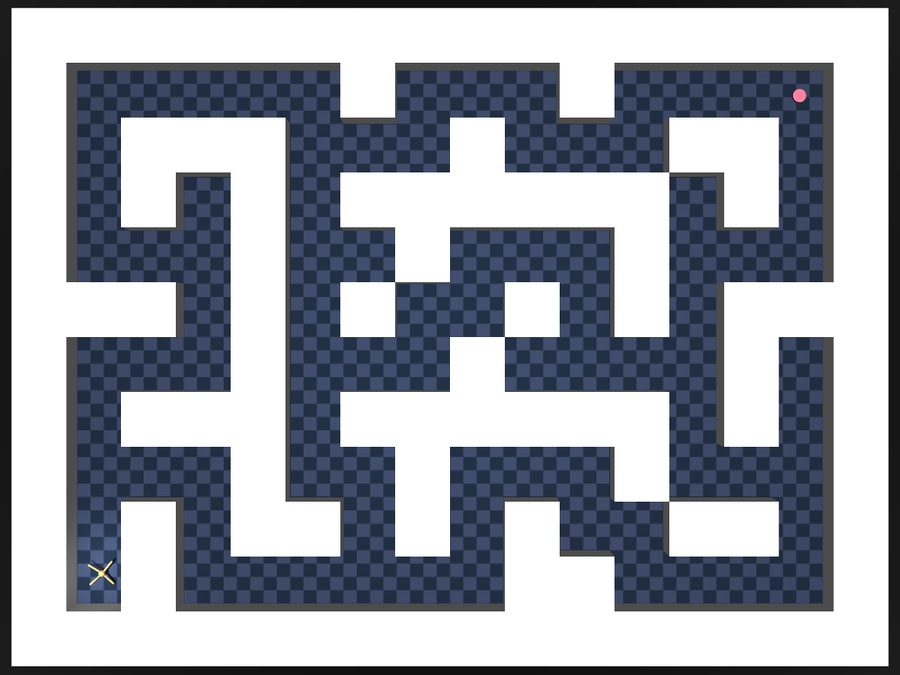} &
    \includegraphics[width=0.32\textwidth]{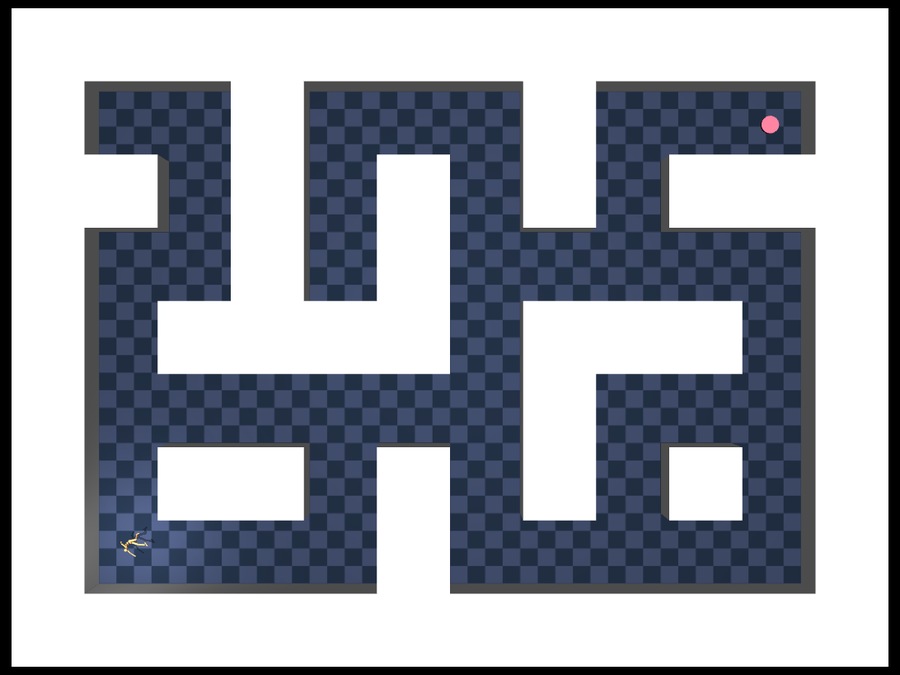} \\
    \texttt{puzzle-4x4} & \texttt{antmaze-giant} & \texttt{humanoidmaze-large}
  \end{tabular}
\caption{The six OGBench domains of our experiments, each shown at the initial state of one of its tasks.}
  \label{fig:domains}
\end{figure}

We evaluate on 30 tasks from six OGBench domains \citep{park2025ogbench}: \texttt{cube-double}, \texttt{cube-triple}, \texttt{cube-quadruple}, \texttt{puzzle-4x4}, \texttt{antmaze-giant} and \texttt{humanoidmaze-large}. Each domain has five single-task variants (\texttt{task1} to \texttt{task5}), which share the dataset of the domain and differ in the goal. Figure~\ref{fig:domains} shows the domains, and Table~\ref{tab:domains} lists them together with the settings that are shared by all methods.

\begin{table}[h]
\caption{Domains and the settings shared by all methods. The discount is per environment step.}
\label{tab:domains}
\begin{center}
\setlength{\tabcolsep}{4.5pt}
\begin{tabular}{lcccccc}
\toprule
Domain & Dataset size & Episode length & Action dim. & Chunk length & Discount & Reward \\
\midrule
\texttt{cube-double}        & 1M   & 500  & 5  & 5 & 0.99  & dense \\
\texttt{cube-triple}        & 3M   & 1000 & 5  & 5 & 0.99  & dense \\
\texttt{cube-quadruple}     & 25M  & 1000 & 5  & 5 & 0.99  & dense \\
\texttt{puzzle-4x4}         & 100M & 500  & 5  & 5 & 0.99  & sparse \\
\texttt{antmaze-giant}      & 1M   & 1000 & 8  & 1 & 0.995 & sparse \\
\texttt{humanoidmaze-large} & 2M   & 2000 & 21 & 1 & 0.999 & sparse \\
\bottomrule
\end{tabular}
\end{center}
\end{table}

\textbf{Domain settings.} The datasets are the standard OGBench datasets (\texttt{play} for manipulation and \texttt{navigate} for locomotion), except on \texttt{puzzle-4x4} and \texttt{cube-quadruple}. There we start from the 100M-transition datasets \citep{park2025horizonreduction} used by QAM \citep{li2026qam}, which are stored in 100 shards of 1M transitions, and \texttt{cube-quadruple} uses the first 25 shards.
The chunk lengths follow QC \citep{li2025qchunking} and QAM. The discount of \texttt{antmaze-giant} follows QAM, and that of \texttt{humanoidmaze-large} follows TRQAM \citep{dong2026trustregionqadjoint}. The dense reward is the single-task reward of OGBench, $r = - n_{\mathrm{rem}}$, where $n_{\mathrm{rem}}$ is the number of cubes that are not at their goal. The sparse reward is $-1$ while the task is not solved and $0$ otherwise. This is the default reward of the two locomotion domains, and \texttt{puzzle-4x4} uses it as in QAM.

\textbf{Training.} The training and evaluation protocol is identical to that of QC \citep{li2025qchunking}. Every method is trained for 1M gradient steps on the offline dataset and then for 1M environment steps online, with one gradient step per environment step. The replay buffer of the online phase is initialized with the offline dataset. Its capacity is the larger of 2M transitions and the size of the dataset. When the buffer is full, new transitions overwrite the oldest ones. On the two sharded datasets, the offline phase loads a new shard every 1000 gradient steps, and the online buffer is initialized with the first shard.

\textbf{Evaluation.} Every 100k steps we run 50 evaluation episodes and record the success rate. \method acts with a single action sample in data collection, in the TD target and at evaluation. Each baseline keeps the action selection of its implementation (Appendix~\ref{app:baselines}). All methods are run with the same 8 seeds on every task. All confidence intervals are 95\% bootstrap intervals over the seeds, computed from 5000 resamples. For results aggregated over several tasks, we average the per-task means and resample the seeds within each task.

\textbf{Hyperparameter selection.} The settings of Table~\ref{tab:domains} are fixed for every method. The hyperparameters of each method are selected per domain, on one representative task, and then used on all five tasks. Appendix~\ref{app:hyperparameters} lists the selected values and the sweep ranges.

\textbf{Compute.} Our implementation is written in JAX \citep{jax2018github} and built on the official codebase of QC \citep{li2025qchunking} (\url{https://github.com/ColinQiyangLi/qc}). All experiments are run on NVIDIA RTX 4090 GPUs.
 \FloatBarrier
\section{Baselines}
\label{app:baselines}

We compare \method with four one-step baselines: MVP, FMQ, DFP and \vgname. 
This section states their objectives in the notation of Sections~\ref{sec:prelim} and~\ref{sec:method}. All of them are trained under the protocol of Appendix~\ref{app:benchmark}.

\textbf{Best-of-$N$ selection.} MVP, FMQ and DFP act through selection. 
Let $\pi_\theta(\cdot \mid s)$ denote the distribution of the one-step action of the policy. 
For a state $s$, the policy draws $N$ candidate actions and selects the one with the highest critic value,
\begin{equation}
  a^\star_N(s) = \argmax_{a \in \{a^{(1)}, \dots, a^{(N)}\}} Q(s, a), \qquad a^{(i)} \sim \pi_\theta(\cdot \mid s) .
  \label{eq:bon}
\end{equation}
These methods execute $a^\star_N(s)$ in data collection and at evaluation, and they train the critic with
\begin{equation}
  \gL_{\mathrm{TD}}(\phi) = \E_{(s, a, r, s') \sim \gD} \Big( Q_\phi(s, a) - r - \gamma\, Q_{\bar{\phi}} \big( s', a^\star_N(s') \big) \Big)^2 ,
  \label{eq:td_bon}
\end{equation}
 FMQ and MVP use $N = 32$, and DFP uses $N = 16$. \method and \vgname use a single sample in all three places.

\textbf{MVP} \citep{zhan2026meanflowpolicy} trains a flow-map policy $X^\theta$ to imitate the actions in the replay buffer. Its objective consists of the two losses of Section~\ref{sec:behavior}, applied to the policy,
\begin{equation}
  \gL_{\mathrm{MVP}}(\theta) = \gL_{\mathrm{cons}}(\theta) + \gL_{\mathrm{diag}}(\theta) ,
  \label{eq:mvp}
\end{equation}
where $\gL_{\mathrm{cons}}$ is the MeanFlow loss of \eqref{eq:cons} and $\gL_{\mathrm{diag}}$ is the conditional flow matching loss on the diagonal. 
The objective is the same in the offline and in the online phase. The critic enters only through the selection of \eqref{eq:bon}, since the selected actions are stored in $\gD$ and imitated. We use the reproduction of MVP from the FMQ implementation.

\textbf{FMQ} \citep{ziakas2026aligningflowmappolicies} trains a flow-map policy with \eqref{eq:mvp} in the offline phase. At the start of the online phase, the offline policy is frozen. We denote its average velocity by $v_\beta$ and its flow map by $X^\beta$, and the policy $v_\theta$ continues from the same parameters with the loss
\begin{equation}
  \gL_{\mathrm{FMQ}}(\theta) = \E\, \bigg\| v_\theta(x_t, t, 1) - \mathrm{sg} \bigg[ v_\beta(x_t, t, 1) + \kappa\, \frac{\nabla_a Q(s, a^\beta)}{\| \nabla_a Q(s, a^\beta) \|} \bigg] \bigg\|^2 ,
  \qquad a^\beta = X^\beta_{t,1}(x_t) ,
  \quad \kappa = \frac{1}{1 + c\, \tilde{\sigma}} ,
  \label{eq:fmq}
\end{equation}
where $x_t = (1 - t)\, x_0 + t\, x_1$ interpolates between noise and an action of the replay buffer as in \eqref{eq:cfm}, $\tilde{\sigma}$ is the standard deviation of the two critic values at $(s, a^\beta)$ divided by its batch mean, and $c$ is a coefficient. We run the official implementation without modification (\url{https://github.com/chziakas/q-guided-flow-map-policies}).

\textbf{DFP} \citep{koo2026driftingfieldpolicyonestep} uses a drifting model \citep{deng2026generativemodelingdrifting} as the policy, $a = f_\theta(x_0)$. It is trained with the drifting loss, which regresses a generated action onto itself shifted by a drifting field,
\begin{equation}
  \gL_{\mathrm{drift}}(\theta; p) = \E_{s \sim \gD, \; x_0 \sim \gN(0, I_d)} \Big\| \hat{a} - \mathrm{sg} \big[ \hat{a} + \mathbf{V}_{p}(\hat{a}) - \mathbf{V}_{\pi_\theta}(\hat{a}) \big] \Big\|^2 ,
  \qquad
  \mathbf{V}_{q}(a) = \frac{\E_{y \sim q} \big[ k(a, y)\, (y - a) \big]}{\E_{y \sim q} \big[ k(a, y) \big]} ,
  \label{eq:drift}
\end{equation}
where $\hat{a} = f_\theta(x_0)$ and $p$ is a distribution of positive actions for the state $s$. The kernel decays exponentially with the distance between two actions, \mbox{$k(a, y) = \exp\big(-\|a - y\| / (h\, \bar{\Delta})\big)$} \citep{deng2026generativemodelingdrifting}. Here $h$ is the temperature and $\bar{\Delta}$ is the mean distance between the generated actions and their targets in the batch. The field attracts $\hat{a}$ to the positive actions and repels it from the other actions generated by the policy. In the offline phase, the positive actions are the dataset actions, which gives the behavior cloning loss $\gL_{\mathrm{BC}}(\theta) = \gL_{\mathrm{drift}}(\theta; \gD)$. In the online phase, DFP minimizes
\begin{equation}
  \gL_{\mathrm{DFP}}(\theta) = \gL_{\mathrm{BC}}(\theta) + w\, \gL_{\mathrm{drift}}\big(\theta; P_m\big) ,
  \qquad
  P_m = \operatorname*{arg\,top\text{-}\mathit{m}}_{a \in \{a^{(1)}, \dots, a^{(N)}\}} Q(s, a), \quad a^{(i)} \sim \pi_{\bar{\theta}}(\cdot \mid s) ,
  \label{eq:dfp}
\end{equation}
where $\bar{\theta}$ is an exponential moving average of $\theta$ and $w$ is the weight of the top-$m$ loss. The critic thus enters the objective through selection. We run the official implementation without modification (\url{https://github.com/KAIST-Visual-AI-Group/DFP}).

\textbf{\vgname} (value-gradient Q-learning) is a baseline built on the value-gradient target of WTF \citep{mammadov2026wtfsimulationfreereinforcementlearning}, listed in Table~\ref{tab:targets}. It uses this target in place of the target of \eqref{eq:guide} and is otherwise the same as Algorithm~\ref{alg:bqt}. With the on-policy intermediate action $x_t = \mathrm{sg}[X^\theta_{0,t}(x_0)]$ of \eqref{eq:guide}, its loss is
\begin{equation}
  \gL_{\mathrm{VGQL}}(\theta) = \E_{s \sim \gD, \; x_0 \sim \gN(0, I_d), \; t \sim \gU[0, 1]} \Big\| v_\theta(x_t, t, t) - \mathrm{sg} \Big[ v_\beta(x_t, t, t) + \nabla_{x_t} \big( \lambda\, Q(s, x_1) - \hat{C}_t(x_t) \big) \Big] \Big\|^2 ,
  \label{eq:vgql}
\end{equation}
where $x_\tau = X^\theta_{t,\tau}(x_t)$ and $x_1 = X^\theta_{\tau,1}(x_\tau)$ are two jumps of the policy flow map with $\tau \sim \gU[t, 1]$, and $\hat{C}_t(x_t) = \frac{1 - t}{2} \| v_\theta(x_\tau, \tau, \tau) - v_\beta(x_\tau, \tau, \tau) \|^2$ is the one-sample estimate of the control cost. The gradient is taken through both jumps with $\theta$ held fixed, so the target contains the Jacobian of the policy flow map. The variant \vgname (Euclidean) uses the Euclidean reward gradient of WTF, which replaces the gradient of the term $\lambda\, Q(s, x_1)$ by $\lambda\, \nabla_a Q(s, x_1)$ and leaves the gradient of $\hat{C}_t$ unchanged.
 \FloatBarrier
\section{Hyperparameters}
\label{app:hyperparameters}

Table~\ref{tab:hp_common} lists the hyperparameters of \method that are shared by all domains. \vgname, \vgname (Euclidean) and identity transport are run in the same implementation with the same values. The exception is the number of jumps $K$: as in WTF, the two \vgname variants draw the intermediate time $\tau$ of \eqref{eq:vgql} at random instead. Following QAM \citep{li2026qam}, the target of \eqref{eq:guide} is computed with target networks of the behavior velocity and of the critic. FMQ, MVP and DFP use the hyperparameters of their official implementations. Table~\ref{tab:hp_domain} lists the domain-specific hyperparameters and Table~\ref{tab:hp_sweep} their sweep ranges.

\begin{table}[h]
\caption{Hyperparameters of \method shared by all domains.}
\label{tab:hp_common}
\begin{center}
\begin{tabular}{ll}
\toprule
Hyperparameter & Value \\
\midrule
Optimizer & Adam \\
Learning rate & $3 \times 10^{-4}$ \\
Batch size & 256 \\
Velocity networks $v_\theta$, $v_\beta$ & MLP, 4 hidden layers of width 512, GELU \\
Time embedding & Fourier features of dimension 64, then an MLP of width 128 \\
Critic network & MLP, 4 hidden layers of width 512, GELU, layer norm \\
Number of critics & 2, aggregated by their mean \\
Target network update rate & 0.005 \\
Number of jumps $K$ & 2, evenly spaced \\
Lower limit $\lambda_{\min}$ & 0.02 \\
Clipping of $\hat{\gC} / \delta - 1$ in \eqref{eq:dual} & at most 2 \\
Offline gradient steps & 1M \\
Online environment steps & 1M \\
Update-to-data ratio & 1 \\
Online steps before the first update & 5000 \\
\bottomrule
\end{tabular}
\end{center}
\end{table}

\begin{table}[h]
\caption{Domain-specific hyperparameters. $\delta$ is the control cost budget, $\lambda_{\mathrm{init}}$ the initial value and $\lambda_{\max}$ the upper limit of $\lambda$, and $\eta$ the step size of \eqref{eq:dual}; $a \to b$ denotes the value $a$ in the offline phase and $b$ in the online phase. $c$ is the coefficient of FMQ in \eqref{eq:fmq}, $m$ the number of positive actions of DFP in \eqref{eq:dfp} and $h$ the temperature of its kernel in \eqref{eq:drift}. Identity transport uses the values of \method. $\dagger$: \vgname uses $1.5 \to 4$.}
\label{tab:hp_domain}
\begin{center}
\small
\setlength{\tabcolsep}{4pt}
\begin{tabular}{lccccccccc}
\toprule
 & \multicolumn{3}{c}{$\delta$} & & & & FMQ & \multicolumn{2}{c}{DFP} \\
\cmidrule(lr){2-4} \cmidrule(lr){9-10}
Domain & \method & \vgname & \vgname (Eucl.) & $\lambda_{\mathrm{init}}$ & $\lambda_{\max}$ & $\eta$ $(10^{-5})$ & $c$ & $m$ & $h$ \\
\midrule
\texttt{cube-double}        & 0.3   & 0.9   & --    & 0.05 & 1.5 & 2 & 0.3 & 4 & 0.1 \\
\texttt{cube-triple}        & 0.9   & 0.9   & 0.9   & 0.05 & 1.5 & 2 & 0.3 & 4 & 0.1 \\
\texttt{cube-quadruple}     & 1.8   & 1.8   & 1.8   & 0.05 & 1.5 & 2 & 0.3 & 4 & 0.1 \\
\texttt{puzzle-4x4}         & 3     & 3     & --    & 1    & 100 & 100 & 0.3 & 2 & 0.1 \\
\texttt{antmaze-giant}      & 0.288 & 0.288 & 0.288 & 0.05 & $1.5 \to 1^{\dagger}$ & 2 & 0.3 & 8 & 0.1 \\
\texttt{humanoidmaze-large} & 0.195 & 0.195 & 0.13  & 0.05 & 1.5 & $2 \to 20$ & 10 & 4 & 1 \\
\bottomrule
\end{tabular}
\end{center}
\end{table}

\begin{table}[h]
\caption{Sweep ranges of the domain-specific hyperparameters on the selection task of each domain. The selected value is in bold. Hyperparameters and domains that are not listed use the value of Table~\ref{tab:hp_domain} without a sweep. For FMQ and DFP, this is the value of the official implementation. For \vgname on \texttt{cube-double} and \texttt{puzzle-4x4}, it is the only value we ran, as \vgname already solves the selection task with it. For DFP, $m$ is swept at $h = 0.1$ and $h$ at the selected $m$, and $\{0.1, 0.5\}$ denotes a drifting field summed over the two temperatures.}
\label{tab:hp_sweep}
\begin{center}
\small
\begin{tabular}{llll}
\toprule
Method & Domain & Task & Values \\
\midrule
\method & \texttt{cube-double} & \texttt{task4} & $\delta$: \textbf{0.3}, 0.45, 0.9 \\
 & \texttt{cube-triple} & \texttt{task4} & $\delta$: 0.45, \textbf{0.9}, 1.35 \\
 & \texttt{cube-quadruple} & \texttt{task3} & $\delta$: 0.45, 0.9, \textbf{1.8}, 2.7, 3.6 \\
 & \texttt{puzzle-4x4} & \texttt{task4} & $\delta$: 1.5, \textbf{3}, 6 \\
 & \texttt{antmaze-giant} & \texttt{task3} & $\delta$: 0.144, \textbf{0.288}, 0.576; online $\lambda_{\max}$: \textbf{1}, 1.5, 4 \\
 & \texttt{humanoidmaze-large} & \texttt{task5} & $\delta$: 0.13, \textbf{0.195}, 0.26; online $\eta$ $(10^{-5})$: 2, \textbf{20} \\
\midrule
\vgname & \texttt{cube-triple} & \texttt{task4} & $\delta$: 0.45, \textbf{0.9}, 1.8 \\
 & \texttt{cube-quadruple} & \texttt{task3} & $\delta$: 0.45, 0.9, \textbf{1.8}, 2.7, 3.6 \\
 & \texttt{antmaze-giant} & \texttt{task3} & $\delta$: 0.144, \textbf{0.288}, 0.576; online $\lambda_{\max}$: 1, \textbf{4} \\
 & \texttt{humanoidmaze-large} & \texttt{task5} & $\delta$: 0.13, \textbf{0.195}; online $\eta$ $(10^{-5})$: 2, \textbf{20} \\
\midrule
\vgname (Euclidean) & \texttt{cube-triple} & \texttt{task4} & $\delta$: 0.45, \textbf{0.9}, 1.8 \\
 & \texttt{cube-quadruple} & \texttt{task3} & $\delta$: 0.45, 0.9, \textbf{1.8}, 2.7, 3.6 \\
 & \texttt{antmaze-giant} & \texttt{task3} & $\delta$: 0.144, \textbf{0.288}, 0.576; online $\lambda_{\max}$: \textbf{1}, 4 \\
 & \texttt{humanoidmaze-large} & \texttt{task5} & $\delta$: \textbf{0.13}, 0.195; online $\eta$ $(10^{-5})$: 2, \textbf{20} \\
\midrule
FMQ & \texttt{cube-quadruple} & \texttt{task3} & $c$: \textbf{0.3}, 1, 3 \\
 & \texttt{antmaze-giant} & \texttt{task3} & $c$: 0.1, \textbf{0.3}, 1, 3 \\
 & \texttt{humanoidmaze-large} & \texttt{task5} & $c$: 0.3, 1, 3, \textbf{10}, 30 \\
\midrule
DFP & \texttt{puzzle-4x4} & \texttt{task4} & $m$: 1, \textbf{2}, 4, 8 \\
 & \texttt{antmaze-giant} & \texttt{task3} & $m$: 1, 2, 4, \textbf{8}; $h$: \textbf{0.1}, 0.5 \\
 & \texttt{humanoidmaze-large} & \texttt{task5} & $m$: 1, 2, \textbf{4}, 8; $h$: 0.1, 0.25, 0.5, \textbf{1}, 2, $\{0.1, 0.5\}$ \\
\bottomrule
\end{tabular}
\end{center}
\end{table}
 \FloatBarrier
\section{Compute Cost of Best-of-$N$ Selection}
\label{app:compute}

Table~\ref{tab:flops} quantifies the compute cost of best-of-$N$ selection on the networks of \method. Following \citet{zhou2026testtimegradientguidance} and \citet{zhou2026advantageguidance}, we report the floating-point operations (FLOPs) of one policy query, which returns an action or an action chunk.

With $N = 1$, which is how \method acts in all our experiments, a query costs one forward pass of the policy. With $N > 1$, the policy generates $N$ candidate actions, every critic of the ensemble evaluates each of them, and the candidate with the highest value is executed. We count two FLOPs per weight of every dense layer and average the count over the six domains, whose observation and action dimensions differ. The first configuration is that of the main experiments (Table~\ref{tab:hp_common}), and the second is the configuration with networks of width 1024 and 10 critics of Appendix~\ref{app:multistep}.

\begin{table}[h]
\caption{\textbf{Compute cost of best-of-$N$ selection.} FLOPs of one policy query with the networks of \method, when the action is selected among $N$ samples by the critic ensemble. \method uses $N = 1$.}
\label{tab:flops}
\begin{center}
\small
\begin{tabular}{llrrrrr}
\toprule
Networks & & $N = 1$ & $N = 4$ & $N = 8$ & $N = 16$ & $N = 32$ \\
\midrule
Width 512, 2 critics & FLOPs ($10^6$) & 1.9 & 20.7 & 41.4 & 82.7 & 165.4 \\
 & relative to $N = 1$ & 1$\times$ & 11$\times$ & 22$\times$ & 44$\times$ & 89$\times$ \\
\midrule
Width 1024, 10 critics & FLOPs ($10^6$) & 6.8 & 285.1 & 570.2 & 1140.5 & 2281.0 \\
 & relative to $N = 1$ & 1$\times$ & 42$\times$ & 84$\times$ & 167$\times$ & 335$\times$ \\
\bottomrule
\end{tabular}
 \end{center}
\end{table}

A single critic costs about as much as the policy, so selection among $N$ samples with two critics costs close to $3N$ times a single sample, and the factor grows with the size of the ensemble.
 \FloatBarrier
\section{Main Results per Domain and per Task}
\label{app:tasks}

\begin{figure}[h]
  \centering
  \includegraphics[width=\textwidth]{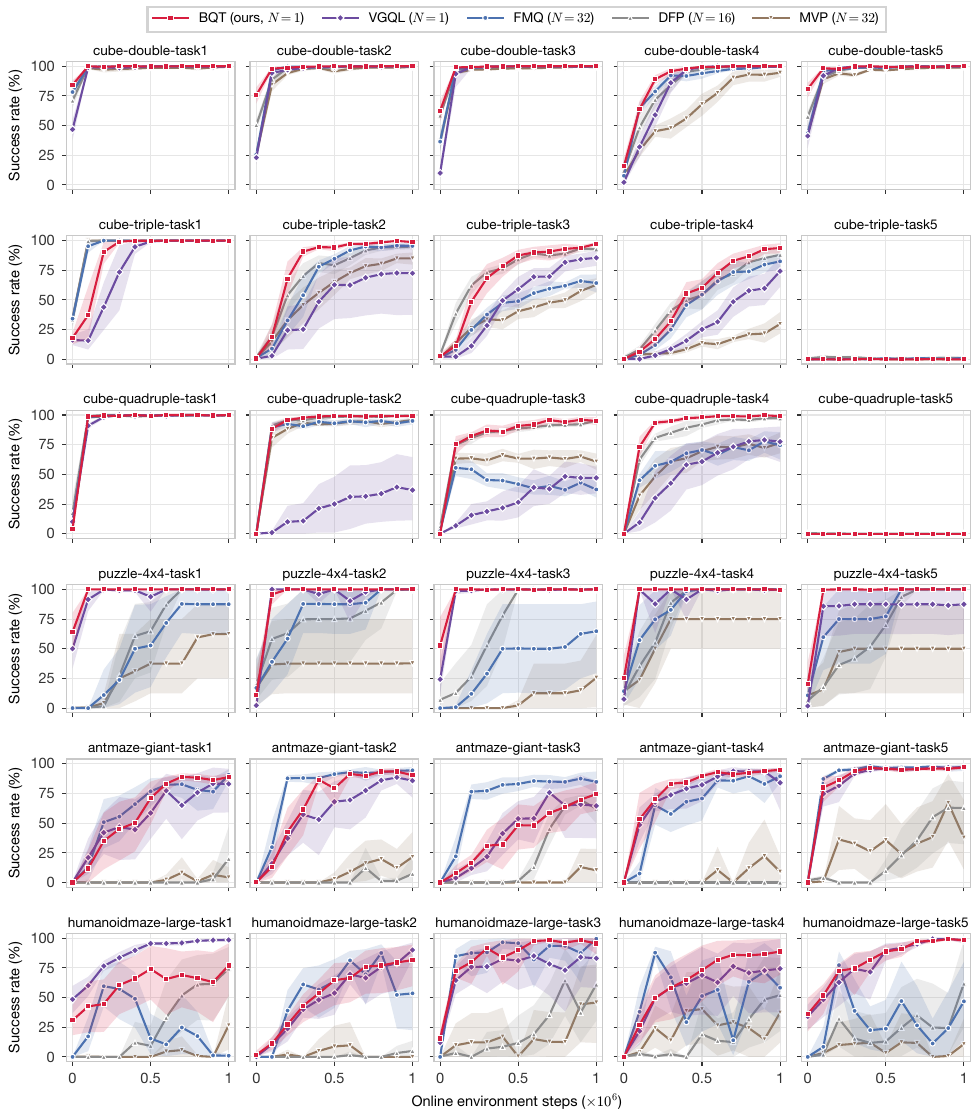}
  \caption{\textbf{Main results per task (8 seeds).} Online phase; step 0 is the offline policy. Means with 95\% bootstrap confidence intervals.}
  \label{fig:main_tasks}
\end{figure}

Table~\ref{tab:main} reports the success rates of the main experiments per domain and Table~\ref{tab:main_tasks} per task. Figure~\ref{fig:main_tasks} shows the training curves of every task.

\begin{table}[h]
\caption{\textbf{Main results, final success rate (\%).} Means over 8 seeds at 2M steps with 95\% bootstrap confidence intervals. Bold: at or above 95\% of the best value in the row.}
\label{tab:main}
\begin{center}
\small
\setlength{\tabcolsep}{4pt}
\begin{tabular}{lccccc}
\toprule
Domain & MVP & DFP & FMQ & \vgname & \method (ours) \\
 & ($N = 32$) & ($N = 16$) & ($N = 32$) & ($N = 1$) & ($N = 1$) \\
\midrule
\texttt{cube-double} & \textbf{98.3} {\tiny [97.5, 99.0]} & \textbf{99.9} {\tiny [99.8, 100.0]} & \textbf{99.8} {\tiny [99.5, 100.0]} & \textbf{99.9} {\tiny [99.8, 100.0]} & \textbf{100.0} {\tiny [100.0, 100.0]} \\
\texttt{cube-triple} & 55.5 {\tiny [53.1, 58.0]} & \textbf{75.3} {\tiny [74.2, 76.5]} & 68.7 {\tiny [66.5, 70.7]} & 66.5 {\tiny [59.7, 72.2]} & \textbf{78.0} {\tiny [77.0, 78.8]} \\
\texttt{cube-quadruple} & 66.5 {\tiny [64.2, 68.7]} & \textbf{78.3} {\tiny [78.0, 78.7]} & 61.4 {\tiny [58.5, 64.0]} & 52.2 {\tiny [45.6, 58.9]} & \textbf{78.7} {\tiny [78.0, 79.3]} \\
\texttt{puzzle-4x4} & 50.1 {\tiny [35.2, 65.0]} & \textbf{100.0} {\tiny [100.0, 100.0]} & 90.5 {\tiny [82.5, 97.5]} & \textbf{97.5} {\tiny [92.5, 100.0]} & \textbf{100.0} {\tiny [99.8, 100.0]} \\
\texttt{antmaze-giant} & 16.6 {\tiny [8.6, 25.0]} & 30.4 {\tiny [22.7, 38.1]} & \textbf{91.4} {\tiny [88.8, 93.8]} & 82.8 {\tiny [74.0, 90.2]} & \textbf{89.1} {\tiny [86.0, 91.5]} \\
\texttt{humanoidmaze-large} & 24.1 {\tiny [13.2, 35.2]} & 50.9 {\tiny [39.2, 62.1]} & 51.8 {\tiny [41.4, 61.5]} & \textbf{88.8} {\tiny [80.5, 95.4]} & \textbf{88.2} {\tiny [80.2, 94.7]} \\
\midrule
All 30 tasks & 51.8 {\tiny [48.6, 55.2]} & 72.5 {\tiny [70.1, 74.8]} & 77.2 {\tiny [74.9, 79.5]} & 81.3 {\tiny [78.6, 83.7]} & \textbf{89.0} {\tiny [87.5, 90.2]} \\
\bottomrule
\end{tabular}
 \end{center}
\end{table}

\begin{table}[h]
\caption{\textbf{Main results per task, final success rate (\%).} Means over 8 seeds at 2M steps with 95\% bootstrap confidence intervals. Bold: at or above 95\% of the best value in the row.}
\label{tab:main_tasks}
\begin{center}
\small
\setlength{\tabcolsep}{4pt}
\begin{tabular}{lccccc}
\toprule
Task & MVP & DFP & FMQ & \vgname & \method (ours) \\
 & ($N = 32$) & ($N = 16$) & ($N = 32$) & ($N = 1$) & ($N = 1$) \\
\midrule
\multicolumn{6}{l}{\texttt{cube-double}} \\
\quad\texttt{task1} & \textbf{99.5} {\tiny [99.0, 100.0]} & \textbf{100.0} {\tiny [100.0, 100.0]} & \textbf{100.0} {\tiny [100.0, 100.0]} & \textbf{99.8} {\tiny [99.2, 100.0]} & \textbf{100.0} {\tiny [100.0, 100.0]} \\
\quad\texttt{task2} & \textbf{99.5} {\tiny [98.8, 100.0]} & \textbf{100.0} {\tiny [100.0, 100.0]} & \textbf{99.5} {\tiny [98.8, 100.0]} & \textbf{99.8} {\tiny [99.2, 100.0]} & \textbf{100.0} {\tiny [100.0, 100.0]} \\
\quad\texttt{task3} & \textbf{99.0} {\tiny [98.0, 99.8]} & \textbf{100.0} {\tiny [100.0, 100.0]} & \textbf{99.8} {\tiny [99.2, 100.0]} & \textbf{100.0} {\tiny [100.0, 100.0]} & \textbf{100.0} {\tiny [100.0, 100.0]} \\
\quad\texttt{task4} & 94.5 {\tiny [91.0, 97.2]} & \textbf{99.5} {\tiny [99.0, 100.0]} & \textbf{99.8} {\tiny [99.2, 100.0]} & \textbf{100.0} {\tiny [100.0, 100.0]} & \textbf{100.0} {\tiny [100.0, 100.0]} \\
\quad\texttt{task5} & \textbf{99.0} {\tiny [98.0, 99.8]} & \textbf{100.0} {\tiny [100.0, 100.0]} & \textbf{99.8} {\tiny [99.2, 100.0]} & \textbf{100.0} {\tiny [100.0, 100.0]} & \textbf{100.0} {\tiny [100.0, 100.0]} \\
\midrule
\multicolumn{6}{l}{\texttt{cube-triple}} \\
\quad\texttt{task1} & \textbf{100.0} {\tiny [100.0, 100.0]} & \textbf{100.0} {\tiny [100.0, 100.0]} & \textbf{100.0} {\tiny [100.0, 100.0]} & \textbf{100.0} {\tiny [100.0, 100.0]} & \textbf{100.0} {\tiny [100.0, 100.0]} \\
\quad\texttt{task2} & 85.0 {\tiny [79.5, 90.0]} & \textbf{95.0} {\tiny [92.8, 97.2]} & \textbf{95.5} {\tiny [93.5, 97.8]} & 72.5 {\tiny [37.5, 97.0]} & \textbf{98.8} {\tiny [97.8, 99.8]} \\
\quad\texttt{task3} & 62.8 {\tiny [56.5, 68.2]} & \textbf{92.8} {\tiny [91.2, 94.5]} & 64.2 {\tiny [56.5, 71.2]} & 85.5 {\tiny [78.8, 91.5]} & \textbf{97.2} {\tiny [95.8, 98.5]} \\
\quad\texttt{task4} & 29.8 {\tiny [21.8, 39.5]} & 88.0 {\tiny [83.0, 93.0]} & 82.5 {\tiny [75.5, 89.2]} & 74.2 {\tiny [60.8, 85.8]} & \textbf{93.8} {\tiny [89.0, 98.0]} \\
\quad\texttt{task5} & 0.0 {\tiny [0.0, 0.0]} & 1.0 {\tiny [0.0, 2.0]} & \textbf{1.2} {\tiny [0.0, 2.8]} & 0.0 {\tiny [0.0, 0.0]} & 0.0 {\tiny [0.0, 0.0]} \\
\midrule
\multicolumn{6}{l}{\texttt{cube-quadruple}} \\
\quad\texttt{task1} & \textbf{99.2} {\tiny [98.5, 100.0]} & \textbf{100.0} {\tiny [100.0, 100.0]} & \textbf{100.0} {\tiny [100.0, 100.0]} & \textbf{99.8} {\tiny [99.2, 100.0]} & \textbf{100.0} {\tiny [100.0, 100.0]} \\
\quad\texttt{task2} & \textbf{95.8} {\tiny [94.2, 97.2]} & \textbf{99.2} {\tiny [98.2, 100.0]} & \textbf{95.0} {\tiny [92.0, 97.5]} & 36.8 {\tiny [12.0, 65.2]} & \textbf{99.2} {\tiny [98.2, 100.0]} \\
\quad\texttt{task3} & 60.8 {\tiny [53.8, 67.8]} & \textbf{95.2} {\tiny [94.1, 96.2]} & 37.2 {\tiny [30.5, 43.8]} & 47.0 {\tiny [34.0, 56.5]} & \textbf{95.0} {\tiny [91.8, 97.8]} \\
\quad\texttt{task4} & 76.5 {\tiny [67.5, 85.2]} & \textbf{97.2} {\tiny [96.2, 98.2]} & 74.5 {\tiny [62.5, 85.8]} & 77.5 {\tiny [60.5, 90.2]} & \textbf{99.0} {\tiny [98.0, 99.8]} \\
\quad\texttt{task5} & 0.0 {\tiny [0.0, 0.0]} & 0.0 {\tiny [0.0, 0.0]} & 0.0 {\tiny [0.0, 0.0]} & 0.0 {\tiny [0.0, 0.0]} & 0.0 {\tiny [0.0, 0.0]} \\
\midrule
\multicolumn{6}{l}{\texttt{puzzle-4x4}} \\
\quad\texttt{task1} & 62.5 {\tiny [25.0, 87.5]} & \textbf{100.0} {\tiny [100.0, 100.0]} & 87.5 {\tiny [62.5, 100.0]} & \textbf{100.0} {\tiny [100.0, 100.0]} & \textbf{100.0} {\tiny [100.0, 100.0]} \\
\quad\texttt{task2} & 37.8 {\tiny [12.5, 75.0]} & \textbf{100.0} {\tiny [100.0, 100.0]} & \textbf{100.0} {\tiny [100.0, 100.0]} & \textbf{100.0} {\tiny [100.0, 100.0]} & \textbf{100.0} {\tiny [100.0, 100.0]} \\
\quad\texttt{task3} & 25.5 {\tiny [0.8, 51.8]} & \textbf{100.0} {\tiny [100.0, 100.0]} & 64.8 {\tiny [29.5, 89.8]} & \textbf{100.0} {\tiny [100.0, 100.0]} & \textbf{100.0} {\tiny [100.0, 100.0]} \\
\quad\texttt{task4} & 75.0 {\tiny [37.5, 100.0]} & \textbf{100.0} {\tiny [100.0, 100.0]} & \textbf{100.0} {\tiny [100.0, 100.0]} & \textbf{100.0} {\tiny [100.0, 100.0]} & \textbf{99.8} {\tiny [99.2, 100.0]} \\
\quad\texttt{task5} & 50.0 {\tiny [12.5, 87.5]} & \textbf{100.0} {\tiny [100.0, 100.0]} & \textbf{100.0} {\tiny [100.0, 100.0]} & 87.5 {\tiny [62.5, 100.0]} & \textbf{100.0} {\tiny [100.0, 100.0]} \\
\midrule
\multicolumn{6}{l}{\texttt{antmaze-giant}} \\
\quad\texttt{task1} & 4.2 {\tiny [0.0, 12.8]} & 20.0 {\tiny [0.8, 47.5]} & \textbf{90.8} {\tiny [85.5, 95.2]} & 83.0 {\tiny [59.0, 95.8]} & \textbf{88.8} {\tiny [85.5, 91.8]} \\
\quad\texttt{task2} & 21.5 {\tiny [5.8, 40.5]} & 7.2 {\tiny [0.0, 21.7]} & \textbf{94.5} {\tiny [91.5, 97.2]} & 85.8 {\tiny [71.5, 94.2]} & \textbf{90.2} {\tiny [86.2, 93.5]} \\
\quad\texttt{task3} & 10.2 {\tiny [0.0, 28.2]} & 62.2 {\tiny [49.8, 73.2]} & \textbf{84.8} {\tiny [74.8, 93.0]} & 64.5 {\tiny [39.2, 85.2]} & 74.5 {\tiny [60.0, 84.8]} \\
\quad\texttt{task4} & 9.0 {\tiny [0.5, 22.5]} & 0.0 {\tiny [0.0, 0.0]} & 89.5 {\tiny [84.0, 94.2]} & 84.0 {\tiny [60.8, 96.5]} & \textbf{95.0} {\tiny [93.5, 96.2]} \\
\quad\texttt{task5} & 37.8 {\tiny [9.5, 69.0]} & 62.8 {\tiny [32.0, 84.8]} & \textbf{97.5} {\tiny [94.0, 99.8]} & \textbf{96.8} {\tiny [95.0, 98.5]} & \textbf{97.0} {\tiny [95.8, 98.2]} \\
\midrule
\multicolumn{6}{l}{\texttt{humanoidmaze-large}} \\
\quad\texttt{task1} & 26.5 {\tiny [6.2, 52.8]} & 74.5 {\tiny [50.0, 92.8]} & 1.0 {\tiny [0.0, 3.0]} & \textbf{98.5} {\tiny [97.0, 99.5]} & 77.0 {\tiny [50.2, 96.0]} \\
\quad\texttt{task2} & 0.0 {\tiny [0.0, 0.0]} & 5.0 {\tiny [0.2, 13.8]} & 53.5 {\tiny [22.7, 82.2]} & \textbf{90.0} {\tiny [82.2, 96.5]} & 81.5 {\tiny [59.8, 94.2]} \\
\quad\texttt{task3} & 45.8 {\tiny [12.5, 78.2]} & 61.0 {\tiny [31.2, 87.5]} & \textbf{99.5} {\tiny [98.5, 100.0]} & 83.0 {\tiny [58.0, 99.5]} & \textbf{95.5} {\tiny [86.5, 100.0]} \\
\quad\texttt{task4} & 37.2 {\tiny [12.0, 74.0]} & 52.0 {\tiny [21.7, 83.0]} & 58.2 {\tiny [29.2, 85.5]} & 74.2 {\tiny [37.5, 99.2]} & \textbf{88.8} {\tiny [69.2, 99.5]} \\
\quad\texttt{task5} & 10.8 {\tiny [0.0, 32.2]} & 61.8 {\tiny [24.8, 87.2]} & 46.8 {\tiny [13.5, 80.0]} & \textbf{98.0} {\tiny [95.8, 99.8]} & \textbf{98.2} {\tiny [95.8, 99.8]} \\
\bottomrule
\end{tabular}
 \end{center}
\end{table}

 \FloatBarrier
\section{Comparison with Multi-Step Flow Policies}
\label{app:multistep}

In the comparison with multi-step flow policies of Section~\ref{sec:results}, \method is run under the protocol of TRQAM \citep{dong2026trustregionqadjoint}. This appendix describes the protocol and lists the settings and the final results (Table~\ref{tab:multistep}).

\textbf{Protocol.} The behavior flow map is first trained alone for 300k gradient steps on $\gL_{\mathrm{diag}}(\beta) + \gL_{\mathrm{cons}}(\beta)$, and the policy is initialized with its parameters. Algorithm~\ref{alg:bqt} then runs for 1M gradient steps on the offline dataset and for 500k environment steps online.
The critic is an ensemble of 10 networks, and the guidance term uses the action gradient of the ensemble mean.
The datasets of \texttt{antmaze-giant} and \texttt{cube-triple} are the 10M-transition datasets of TRQAM, \texttt{cube-quadruple} uses all 100M transitions, and \texttt{humanoidmaze-large} uses the standard dataset.
The replay buffer of the online phase holds the dataset and all online transitions. On the sharded datasets, a new shard is loaded every 1000 steps in both phases.
Every 50k steps we run 50 evaluation episodes, and each task is run with 8 seeds. As in the main experiments, \method acts with a single action sample. The remaining hyperparameters are listed in Tables~\ref{tab:hp_common} and~\ref{tab:hp_multistep}.

\textbf{Baseline results.} The results of TRQAM and QAM-E are those released by the authors of TRQAM (\url{https://github.com/yonghdong/trqam}), with 8 seeds per task. QAM-E is the variant of QAM \citep{li2026qam} with an edit policy \citep{dong2025expo} that modifies the action sampled from the flow policy. Both methods sample an action by integrating the flow with 10 steps.

\begin{table}[h]
\caption{Settings of \method under the protocol of TRQAM. The TD target aggregates the critic ensemble by its mean, minus 0.5 times its standard deviation (std) where stated. The other symbols are those of Table~\ref{tab:hp_domain}, and the values hold for both phases.}
\label{tab:hp_multistep}
\begin{center}
\small
\setlength{\tabcolsep}{4pt}
\begin{tabular}{lcccccccc}
\toprule
Domain & Dataset size & Network width & Discount & TD target & $\delta$ & $\lambda_{\mathrm{init}}$ & $\lambda_{\max}$ & $\eta$ $(10^{-5})$ \\
\midrule
\texttt{cube-triple}        & 10M  & 1024 & 0.995 & mean $-$ 0.5 std & 0.9   & 0.05 & 1.5 & 2 \\
\texttt{cube-quadruple}     & 100M & 1024 & 0.995 & mean $-$ 0.5 std & 1.8   & 0.05 & 1.5 & 2 \\
\texttt{antmaze-giant}      & 10M  & 1024 & 0.995 & mean $-$ 0.5 std & 0.576 & 0.05 & 10  & 2 \\
\texttt{humanoidmaze-large} & 2M   & 512  & 0.999 & mean             & 0.195 & 0.05 & 1.5 & 2 \\
\bottomrule
\end{tabular}
\end{center}
\end{table}

\begin{table}[h]
\caption{\textbf{Comparison with multi-step flow policies, final success rate (\%).} Means over 8 seeds at 1.5M steps with 95\% bootstrap confidence intervals. The results of QAM-E and TRQAM are those released by the authors of TRQAM.}
\label{tab:multistep}
\begin{center}
\small
\begin{tabular}{lccc}
\toprule
Task & QAM-E & TRQAM & \method (ours) \\
\midrule
\texttt{cube-triple-task5} & 91.8 {\tiny [87.0, 96.0]} & 99.2 {\tiny [98.5, 100.0]} & 94.5 {\tiny [90.5, 98.2]} \\
\texttt{cube-quadruple-task5} & 0.5 {\tiny [0.0, 1.5]} & 1.5 {\tiny [0.2, 3.5]} & 0.0 {\tiny [0.0, 0.0]} \\
\texttt{antmaze-giant-task3} & 92.6 {\tiny [90.5, 94.8]} & 94.5 {\tiny [92.5, 96.2]} & 94.8 {\tiny [92.8, 97.0]} \\
\texttt{humanoidmaze-large-task1} & 2.8 {\tiny [0.0, 8.2]} & 77.0 {\tiny [62.2, 87.0]} & 90.5 {\tiny [85.5, 95.0]} \\
\texttt{humanoidmaze-large-task2} & 9.0 {\tiny [0.0, 22.0]} & 36.2 {\tiny [24.5, 44.2]} & 61.0 {\tiny [54.8, 67.0]} \\
\texttt{humanoidmaze-large-task5} & 62.2 {\tiny [25.0, 87.5]} & 81.8 {\tiny [78.0, 86.0]} & 88.0 {\tiny [84.2, 91.2]} \\
\bottomrule
\end{tabular}
 \end{center}
\end{table}
 \FloatBarrier
\section{Additional Ablation Results}
\label{app:ablation}

\begin{figure}[h]
  \centering
  \includegraphics[width=\textwidth]{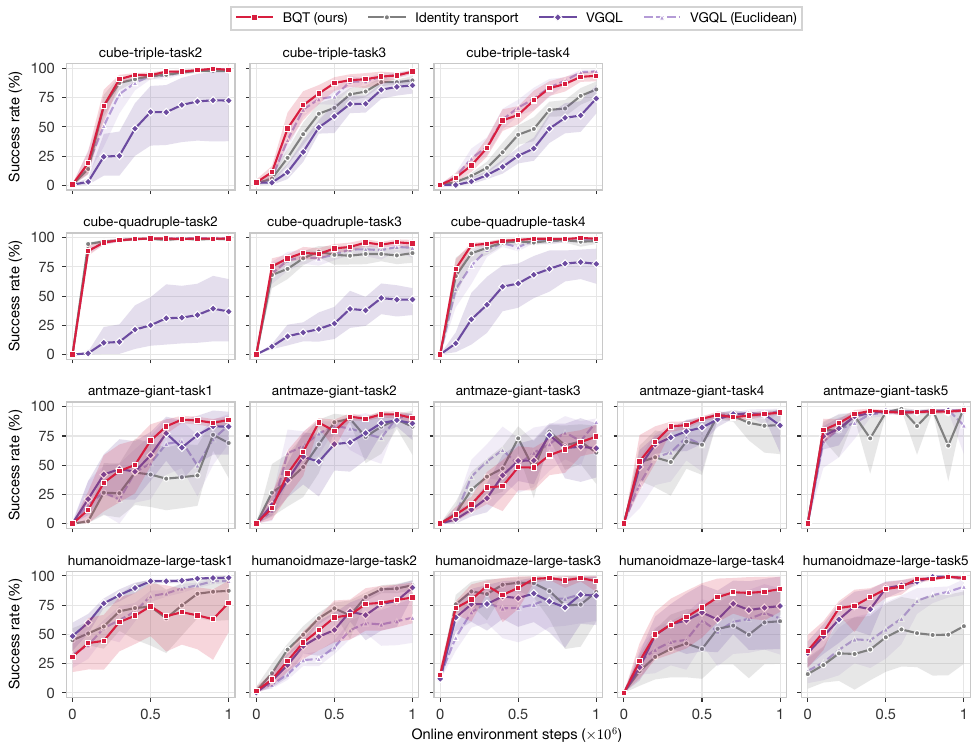}
  \caption{\textbf{Guidance-term ablation per task (8 seeds).} Online phase; step 0 is the offline policy. Means with 95\% bootstrap confidence intervals.}
  \label{fig:abl_target_tasks}
\end{figure}

Figure~\ref{fig:abl_target_tasks} and Table~\ref{tab:abl_target_tasks} give the per-task results of the guidance-term ablation, and Table~\ref{tab:abl_budget} gives the final results of the budget ablation.

\begin{table}[h]
\caption{\textbf{Guidance-term ablation per task, final success rate (\%).} Means over 8 seeds at 2M steps with 95\% bootstrap confidence intervals. Bold: at or above 95\% of the best value in the row.}
\label{tab:abl_target_tasks}
\begin{center}
\small
\setlength{\tabcolsep}{4pt}
\begin{tabular}{lcccc}
\toprule
Task & \vgname & \vgname (Euclidean) & Identity transport & \method (ours) \\
\midrule
\texttt{cube-triple-task2} & 72.5 {\tiny [37.8, 97.2]} & \textbf{99.2} {\tiny [98.5, 99.8]} & \textbf{98.0} {\tiny [96.5, 99.5]} & \textbf{98.8} {\tiny [97.8, 99.8]} \\
\texttt{cube-triple-task3} & 85.5 {\tiny [78.8, 91.2]} & \textbf{96.5} {\tiny [94.0, 99.0]} & 89.8 {\tiny [86.0, 93.5]} & \textbf{97.2} {\tiny [95.8, 98.5]} \\
\texttt{cube-triple-task4} & 74.2 {\tiny [60.8, 85.5]} & \textbf{97.5} {\tiny [96.0, 99.0]} & 82.0 {\tiny [74.2, 88.5]} & \textbf{93.8} {\tiny [89.0, 97.8]} \\
\midrule
\texttt{cube-quadruple-task2} & 36.8 {\tiny [12.0, 64.5]} & \textbf{99.5} {\tiny [98.8, 100.0]} & \textbf{98.5} {\tiny [97.2, 99.5]} & \textbf{99.2} {\tiny [98.2, 100.0]} \\
\texttt{cube-quadruple-task3} & 47.0 {\tiny [34.0, 56.5]} & \textbf{91.2} {\tiny [83.0, 96.8]} & 86.8 {\tiny [78.5, 94.5]} & \textbf{95.0} {\tiny [91.8, 98.0]} \\
\texttt{cube-quadruple-task4} & 77.5 {\tiny [60.2, 90.5]} & \textbf{98.0} {\tiny [96.5, 99.5]} & \textbf{97.0} {\tiny [94.5, 98.8]} & \textbf{99.0} {\tiny [98.0, 99.8]} \\
\midrule
\texttt{antmaze-giant-task1} & 83.0 {\tiny [58.8, 95.8]} & 84.2 {\tiny [68.0, 93.8]} & 69.0 {\tiny [39.0, 90.8]} & \textbf{88.8} {\tiny [85.5, 92.0]} \\
\texttt{antmaze-giant-task2} & 85.8 {\tiny [71.5, 94.2]} & \textbf{92.0} {\tiny [90.0, 94.2]} & 83.0 {\tiny [58.5, 96.2]} & \textbf{90.2} {\tiny [86.2, 93.5]} \\
\texttt{antmaze-giant-task3} & 64.5 {\tiny [39.8, 85.8]} & \textbf{87.0} {\tiny [83.8, 90.0]} & 60.8 {\tiny [35.8, 84.0]} & 74.5 {\tiny [60.0, 84.8]} \\
\texttt{antmaze-giant-task4} & 84.0 {\tiny [60.8, 96.5]} & \textbf{97.0} {\tiny [95.8, 98.5]} & 84.2 {\tiny [59.8, 98.0]} & \textbf{95.0} {\tiny [93.5, 96.2]} \\
\texttt{antmaze-giant-task5} & \textbf{96.8} {\tiny [95.0, 98.2]} & 84.0 {\tiny [59.2, 97.5]} & \textbf{98.8} {\tiny [98.2, 99.5]} & \textbf{97.0} {\tiny [95.8, 98.2]} \\
\midrule
\texttt{humanoidmaze-large-task1} & \textbf{98.5} {\tiny [97.0, 99.8]} & \textbf{95.8} {\tiny [94.2, 97.5]} & 87.2 {\tiny [62.2, 100.0]} & 77.0 {\tiny [51.2, 95.8]} \\
\texttt{humanoidmaze-large-task2} & \textbf{90.0} {\tiny [82.5, 96.5]} & 64.2 {\tiny [42.5, 80.8]} & \textbf{92.0} {\tiny [87.8, 96.2]} & 81.5 {\tiny [59.5, 94.5]} \\
\texttt{humanoidmaze-large-task3} & 83.0 {\tiny [58.0, 99.5]} & 83.5 {\tiny [61.5, 99.5]} & 86.2 {\tiny [61.2, 99.5]} & \textbf{95.5} {\tiny [86.5, 100.0]} \\
\texttt{humanoidmaze-large-task4} & 74.2 {\tiny [37.5, 99.5]} & 63.8 {\tiny [32.8, 89.5]} & 61.2 {\tiny [24.8, 87.0]} & \textbf{88.8} {\tiny [69.2, 99.5]} \\
\texttt{humanoidmaze-large-task5} & \textbf{98.0} {\tiny [95.8, 100.0]} & 90.8 {\tiny [83.8, 97.2]} & 57.0 {\tiny [24.8, 86.8]} & \textbf{98.2} {\tiny [95.8, 99.8]} \\
\bottomrule
\end{tabular}
 \end{center}
\end{table}

\begin{table}[h]
\caption{\textbf{Budget ablation.} $\star$ marks the setting of the main results. Success rates are means over 8 seeds at 2M steps with 95\% bootstrap confidence intervals. Control cost and $\lambda$ are means over the last 500k steps.}
\label{tab:abl_budget}
\begin{center}
\small
\begin{tabular}{llccc}
\toprule
Task & Budget & Success rate (\%) & Control cost / $\delta$ & $\lambda$ \\
\midrule
\texttt{cube-double-task4} & $\delta = 0.3$ $\star$ & 100.0 {\tiny [100.0, 100.0]} & 0.99 & 0.14 \\
 & $\delta = 0.45$ & 100.0 {\tiny [100.0, 100.0]} & 1.00 & 0.20 \\
 & $\delta = 0.9$ & 100.0 {\tiny [100.0, 100.0]} & 0.99 & 0.39 \\
\midrule
\texttt{cube-triple-task4} & $\delta = 0.45$ & 93.5 {\tiny [90.0, 96.5]} & 1.00 & 0.28 \\
 & $\delta = 0.9$ $\star$ & 93.8 {\tiny [89.0, 97.8]} & 1.01 & 0.45 \\
 & $\delta = 1.35$ & 86.0 {\tiny [80.0, 91.5]} & 1.01 & 0.66 \\
\midrule
\texttt{cube-quadruple-task3} & $\delta = 0.9$ & 93.0 {\tiny [89.8, 96.0]} & 0.99 & 0.37 \\
 & $\delta = 1.8$ $\star$ & 95.0 {\tiny [91.8, 98.0]} & 0.98 & 0.67 \\
 & $\delta = 2.7$ & 66.8 {\tiny [50.7, 81.0]} & 0.98 & 1.00 \\
\midrule
\texttt{puzzle-4x4-task4} & $\delta = 1.5$ & 87.5 {\tiny [62.5, 100.0]} & 0.85 & 18.77 \\
 & $\delta = 3$ $\star$ & 99.8 {\tiny [99.2, 100.0]} & 1.00 & 3.09 \\
 & $\delta = 6$ & 100.0 {\tiny [100.0, 100.0]} & 1.01 & 4.47 \\
\midrule
\texttt{humanoidmaze-large-task5} & $\delta = 0.13$ & 86.8 {\tiny [78.8, 92.2]} & 0.99 & 0.29 \\
 & $\delta = 0.195$ $\star$ & 98.2 {\tiny [95.8, 99.8]} & 1.00 & 0.46 \\
 & $\delta = 0.26$ & 99.0 {\tiny [98.0, 99.8]} & 0.99 & 0.63 \\
\midrule
\texttt{antmaze-giant-task3} & $\lambda_{\max} = 1$ $\star$ & 74.5 {\tiny [60.5, 84.5]} & 0.54 & 1.00 \\
 & $\lambda_{\max} = 1.5$ & 49.8 {\tiny [22.7, 75.0]} & 0.73 & 1.49 \\
 & $\lambda_{\max} = 4$ & 45.0 {\tiny [12.8, 77.2]} & 0.97 & 2.15 \\
\bottomrule
\end{tabular}
 \end{center}
\end{table}

\end{document}